\documentclass[journal]{IEEEtran}
\usepackage{cite}

\ifCLASSINFOpdf
  \usepackage[pdftex]{graphicx}
\else
  \usepackage[dvips]{graphicx}
\fi
\usepackage{amsmath}
\usepackage{amssymb}
\usepackage[ruled,vlined]{algorithm2e}

\usepackage{array}
\newcolumntype{L}[1]{>{\raggedright\arraybackslash}m{#1}}
\newcolumntype{C}[1]{>{\centering\arraybackslash}p{#1}}

\usepackage{booktabs}

\ifCLASSOPTIONcompsoc
  \usepackage[caption=false,font=normalsize,labelfont=sf,textfont=sf]{subfig}
\else
  \usepackage[caption=false,font=footnotesize]{subfig}
\fi
\usepackage{url}

\usepackage{bm}
\usepackage{multirow}
\usepackage{multicol}
\usepackage{wrapfig}

\usepackage{balance}

\begin{document}
%
\title{RODNet: A Real-Time Radar Object Detection Network Cross-Supervised by Camera-Radar Fused Object 3D Localization}
%
%
%

\author{Yizhou~Wang,~\IEEEmembership{Student~Member,~IEEE,}
        Zhongyu~Jiang,~\IEEEmembership{Student~Member,~IEEE,}
        Yudong~Li,
        Jenq-Neng~Hwang,~\IEEEmembership{Fellow,~IEEE,}
        Guanbin~Xing,
        and~Hui~Liu,~\IEEEmembership{Fellow,~IEEE}
\thanks{The authors are with the Department
of Electrical and Computer Engineering, University of Washington, Seattle,
WA 98105, USA. E-mail: \texttt{\{ywang26, zyjiang, yudonl, hwang, gxing, huiliu\}@uw.edu}.}
\thanks{H. Liu is also with Silkwave Holdings Limited, Hong Kong.}
\thanks{Manuscript received September 1, 2020; revised December 14, 2020.}}

%
%

\markboth{Journal of Selected Topics in Signal Processing,~Vol.~X, No.~X, March~2021}%
{Y. Wang \MakeLowercase{\textit{et al.}}: RODNet: A Real-Time Radar Object Detection Network Cross-Supervised by Camera-Radar Fused Object 3D Localization}
%



\maketitle

\begin{abstract}
Various autonomous or assisted driving strategies have been facilitated through the accurate and reliable perception of the environment around a vehicle. Among the commonly used sensors, radar has usually been considered as a robust and cost-effective solution even in adverse driving scenarios, e.g., weak/strong lighting or bad weather. Instead of considering to fuse the unreliable information from all available sensors, perception from pure radar data becomes a valuable alternative that is worth exploring. However, unlike rich RGB-based images captured by a camera, it is noticeably difficult to extract semantic information from the radar signals. In this paper, we propose a deep radar object detection network, named RODNet, which is cross-supervised by a camera-radar fused algorithm without laborious annotation efforts, to effectively detect objects from the radio frequency (RF) images in real-time. First, the raw signals captured by millimeter-wave radars are transformed to RF images in range-azimuth coordinates. Second, our proposed RODNet takes a sequence of RF images as the input to predict the likelihood of objects in the radar field of view (FoV). Two customized modules are also added to handle multi-chirp information and object relative motion. Instead of using human-labeled ground truth for training, the proposed RODNet is cross-supervised by a novel 3D localization of detected objects using a camera-radar fusion (CRF) strategy in the training stage. Finally, we propose a method to evaluate the object detection performance of the RODNet. Due to no existing public dataset available for our task, we create a new dataset, named CRUW\footnote{The dataset and code are available at \url{https://www.cruwdataset.org/}.}, which contains synchronized RGB and RF image sequences in various driving scenarios. With intensive experiments, our proposed cross-supervised RODNet achieves 86\% average precision and 88\% average recall of object detection performance, which shows the robustness to noisy scenarios in various driving conditions. 
\end{abstract}


%
\IEEEpeerreviewmaketitle

\section{Introduction}
%
%
%
%

\begin{figure}[!t]
\centering
   \includegraphics[width=0.95\linewidth]{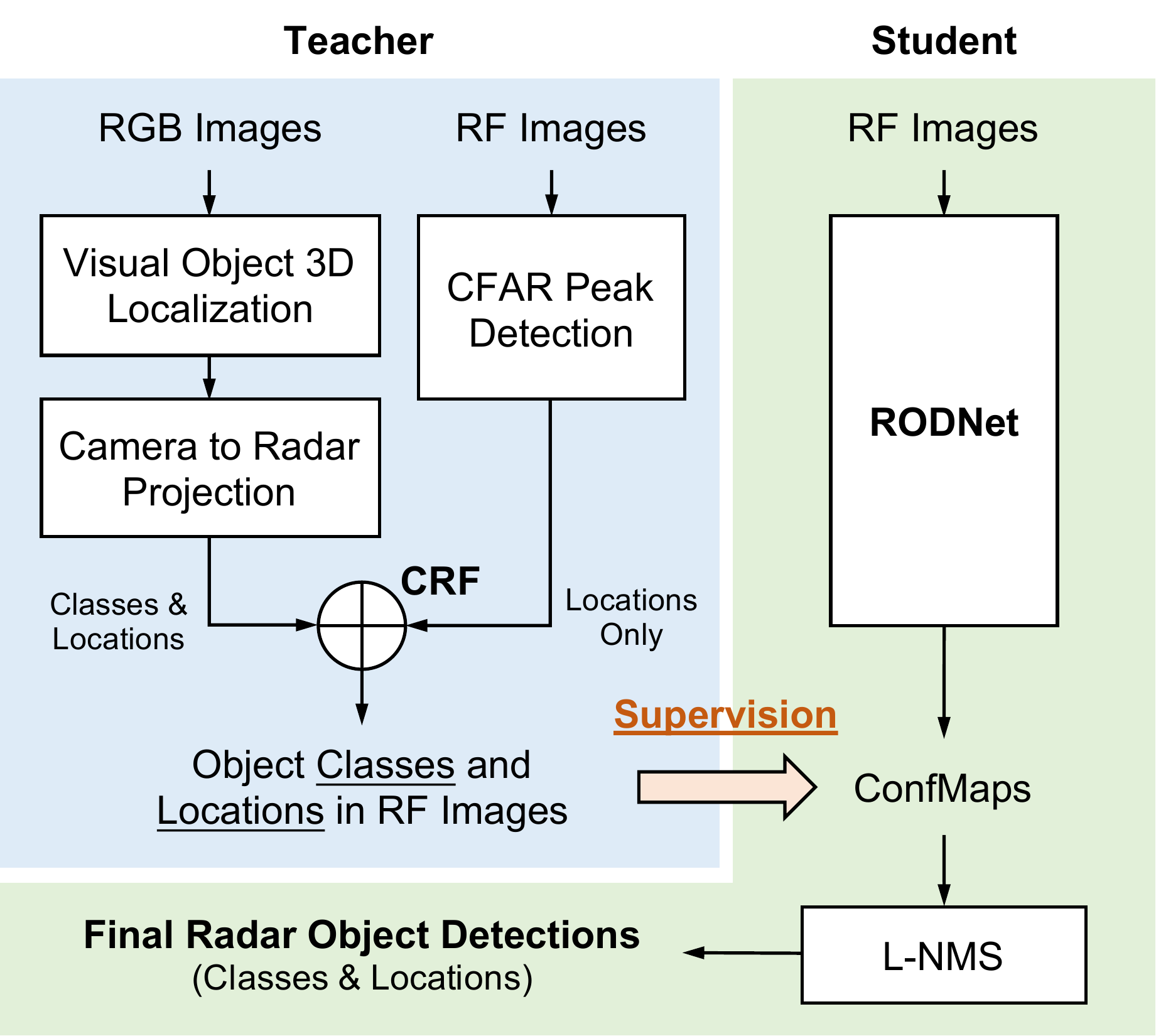}
   \caption{The proposed cross-modal supervision pipeline for radar object detection. Teacher's pipeline fuses the results from both RGB and RF images to obtain the object classes and locations in RF images. Student's pipeline utilizes \textbf{only} RF images as the input to predict the corresponding ConfMaps under the teacher's supervision. The L-NMS as post-processing is followed to calculate the final radar object detection results. }
\label{fig:pipeline}
\end{figure}

\IEEEPARstart{R}{adar} is a common sensor for driving assistance these days, since it is effective in most driving scenarios, including different weather and lighting conditions, resulting in its robustness compared with other sensors, e.g., camera and LiDAR. Many autonomous or assisted driving solutions focus on sensor fusion to improve the accuracy and reliability of the perception results, where radar is mostly used as a complement for cameras or LiDARs. It is mainly because most fusion approaches only take advantage of the more robust localization information in the radar signals, while the rich semantic information hasn't been well exploited. Thus, in this paper, we manage to extract the semantic features of the radio frequency (RF) images by addressing a radar object detection task solely based on radar signals. 

Object detection, which is aimed to detect different objects with their classes and locations, has been a crucial task for many applications. 
Recently, many effective image-based object detectors have been proposed \cite{ren2015faster,lin2017focal,he2017mask,cai2018cascade} and widely used in autonomous or assisted driving systems \cite{mousavian20173d,ansari2018earth,wang2019monocular}. 
Although cameras can give us better semantic understandings of visual scenes, it is not a robust sensor under adverse conditions, such as weak/strong lighting or bad weather, which lead to little/high exposure or blur/occluded images.  
On the other hand, LiDAR is an alternative sensor that can be used for accurate object detection and localization from its point cloud data. After the pioneer research on feature extraction from point cloud \cite{qi2017pointnet, qi2017pointnet++}, subsequent object detection from LiDAR point cloud has been addressed \cite{qi2018frustum, shi2019pointrcnn, shi2020points}. However, these methods require relatively dense LiDAR point cloud for detailed semantic information, not to mention its high equipment and computational costs.

Radar, on the other hand, is relatively more reliable in most harsh environments. Frequency modulated continuous wave (FMCW) radar, which operates in the millimeter-wave (MMW) band (30-300GHz) that is lower than visible light, has the following properties: 1) MMW has great capability to penetrate through fog, smoke, and dust; 2) The huge bandwidth and high working frequency give FMCW radar great range detection ability. 
Typically, there are two kinds of data representations for the FMCW radar, i.e., RF image and radar points. The RF images are generated from the raw radar signals using a series of fast Fourier transforms (FFTs), and the radar points are then derived from these frequency images through a peak detection algorithm \cite{richards2005fundamentals}. Although the radar points can be directly used as the input of the methods designed for the LiDAR point cloud \cite{schumann2018semantic,scheiner2020seeing}, these radar points are usually much sparser (less than 5 points on a nearby car) than the point cloud from a LiDAR \cite{nuscenes2019, feng2020deep}, so that the information is not enough to accomplish the object detection task. 
Whereas, the RF images can maintain the rich Doppler and object motion information so as to have the capability of understanding the semantic meaning of a certain object.

In this paper, we propose a radar object detection method, cross-supervised by a camera-radar fusion algorithm in the training stage, that can accurately detect objects purely based on the radar signals. 
More specifically, we propose a novel radar object detection pipeline, which consists of two parts: a teacher and a student. The teacher's pipeline estimates object classes and 3D locations in the field of view (FoV) by systematically fusing the information from a reliable camera-radar sensor fusion algorithm. The student's pipeline includes a radar object detection network (RODNet) that only takes RF image sequences as the input and estimates object confidence maps (ConfMaps, discussed in Section~\ref{subsec:confmap}). From the ConfMaps, we can further infer the object classes and locations in the radar's range-azimuth coordinates through our post-processing method, called location-based non-maximum suppression (L-NMS, discussed in Section~\ref{subsec:confmap}). The RODNet in the student's pipeline is trained by the annotations systematically labeled by the teacher pipeline without the laborious and unreliable human labeling efforts. 
The aforementioned proposed pipeline is shown in Fig.~\ref{fig:pipeline}. 
As for the network architectures of the RODNet, we implement a 3D convolution neural network (3D CNN) based on an hourglass (HG) architecture with skip connections \cite{newell2016stacked} for feature extraction from RF images. Several customized modules are designed to take advantage of the special properties of the RF image sequences. 
First, chirp information in each radar frame, which contains the detailed object features, is considered. Thus, a chirp merging module (M-Net) is proposed to combine the chirp-level features into the frame-level features. Second, since the radar reflection patterns vary with time due to the relative motion between radar and objects, the classical 3D convolution cannot effectively extract temporal features. Thus, a novel convolution operation, called temporal deformable convolution (TDC), is proposed to handle the temporal evolution of the features in RF image sequences.

We train and evaluate the RODNet using our self-collected dataset, called Camera-Radar of the University of Washington (CRUW), which contains various driving scenarios of about 400K synchronized camera-radar frames. As mentioned above, instead of using radar points as the data format, we choose RF images, which inherently contain detailed motion and surface texture information of objects.
To assess the quantitative performance of our proposed RODNet, without the definition of bounding box widely used in image-based object detection, we further introduce an evaluation method to evaluate the radar object detection performance in RF images. 
With intensive experiments, our RODNet can achieve about $86\%$ AP and $88\%$ AR for object detection performance solely based on RF images in various driving scenarios, regardless of whether objects are visible or not in the camera's FoV.

Overall, our main contributions are the following:
\begin{itemize}
    \item A novel and robust radar object detection network called RODNet for robust object detection in various driving scenarios, which can be used for autonomous or assisted driving without camera or LiDAR information. 
    \item Customized modules, i.e., M-Net and temporal deformable convolution (TDC), are introduced to take advantage of the special properties of RF images effectively.
    \item A camera-radar fusion (CRF) supervision framework for training the RODNet, taking advantage of a monocular camera based object detection and 3D localization method facilitated with statistical detection inference of radar RF images. 
    \item A new dataset, named CRUW, containing synchronized and calibrated camera-radar frames, is collected and can serve as a valuable dataset for camera/radar cross-modal research. 
    \item A new evaluation method for RF image based radar object detection task is proposed and justified for its effectiveness.
\end{itemize}

The rest of this paper is organized as follows. Related works for camera and radar data learning are reviewed in Section~\ref{sec:relatedworks}. The introduction on our proposed RODNet with customized modules is explained in Section~\ref{sec:rod}. The proposed CRF cross-modal supervision framework, obtaining reliable radar object annotations, is addressed in Section~\ref{sec:cross_modal}. In Section~\ref{sec:dataset}, we present our self-collected CRUW dataset used for our training and evaluation. Then, the evaluation method and the experiments are shown in Section~\ref{sec:experiments}. Finally, we conclude our work in Section~\ref{sec:conclusion}.

\section{Related Works}
\label{sec:relatedworks}

\subsection{Learning of Vision Data}

Image-based object detection \cite{girshick2015fast,ren2015faster,he2017mask,cai2018cascade,liu2016ssd,redmon2018yolov3,lin2017focal,duan2019centernet}, which is fundamental and crucial for many computer vision applications, is aimed to detect every object with its class and precise bounding box location from RGB images. 
Given the object detection results, most tracking algorithms focus on exploiting the associations between the detected object bounding boxes in consecutive frames, the so-called tracking-by-detection framework \cite{bergmann2019tracking,yang2019video,tang2019moana,wang2019exploit,cai2020ia,zhanglifts,hsu2020traffic}. Among them, the TrackletNet Tracker (TNT) \cite{wang2019exploit} is an effective and robust tracker to perform multiple object tracking (MOT) of the detected objects with a static or moving camera. Once the same objects among several consecutive frames are associated, the missing and erroneous detections can be recovered or corrected, resulting in better subsequent 3D localization performance. Thus, we implement this tracking technique into the vision part of our framework. 

Object 3D localization has attracted great interest in autonomous and assisted driving communities \cite{song2014robust,song2015joint,mousavian20173d,murthy2017reconstructing,ansari2018earth}.
One idea is to localize vehicles by estimating their 3D structures using a CNN, e.g., 3D bounding boxes \cite{mousavian20173d} and 3D keypoints \cite{murthy2017reconstructing,ansari2018earth,kim2017road}. 
A pre-defined 3D vehicle model is then used to estimate the deformations, resulting in accurate 3D keypoints and the vehicle location. 
Another idea \cite{song2014robust,song2015joint}, however, tries to develop a real-time monocular structure-from-motion (SfM) system, taking into account different kinds of cues, including SfM cues (3D points and ground plane) and object cues (bounding boxes and detection scores). 
Although these works achieve favorable performance in object 3D localization, they only work for the vehicles since only the rigid-body vehicle structure is assumed. 
To overcome this limitation, an accurate and robust object 3D localization system, based on the detected and tracked 2D bounding boxes of objects, is proposed in \cite{wang2019monocular}, which can work for most common moving objects in the road scenes, such as cars, pedestrians, and cyclists. Thus, this monocular camera based 3D localization system is adopted, fused with radar localization information, as our systematic camera annotation method to provide the ground truth for RODNet training.

\subsection{Learning of Radar Data}

Significant research in radar object classification has demonstrated its feasibility as a good alternative when cameras fail to provide good sensing performance \cite{6042174,8468324,capobianco2017vehicle,kwon2017human,cao2018radar,perez2018single,patel2019deep}. With handcrafted feature extraction, Heuel, \textit{et al.}~\cite{6042174} classify radar objects using a support vector machine (SVM) to distinguish cars and pedestrians. Moreover, Angelov \textit{et al.}~\cite{8468324} use a neural network to extract features from the short-time Fourier transform (STFT) heatmap of radar signals. 
However, the above methods only focus on the \emph{classification} tasks, which assume only one object has been appropriately identified in the scene.
Recently, a radar object detection method is proposed in \cite{gao2019experiments}, which combines a statistical constant false alarm rate (CFAR) \cite{richards2005fundamentals} detection algorithm with a CNN-based VGG-16 classifier \cite{simonyan2014very}. 
All of the above approaches are not applicable to the complex driving scenarios with noisy background reflections, e.g., trees, buildings, traffic signs, etc., and could easily give many \textit{false positives}. Besides, the laborious human annotations on the radar RF images required by these methods are usually impossible to obtain. 

Recently, the concept of cross-modal learning has been proposed in the machine learning community \cite{karpathy2015deep,venugopalan2015sequence,qi2016sketch,jing2019self}. This concept is trying to transfer or fuse the information between two different signal modalities to help train the neural networks. 
Specifically, RF-Pose \cite{zhao2018through} introduces the cross-modal supervision idea into wireless signals to achieve human pose estimation based on WiFi range radio signals, using a computer vision technique, i.e., OpenPose \cite{cao2017realtime}, to systematically generate annotations of human body keypoints from the camera. 
However, radar object detection is more challenging: 1) Feature extraction for object detection (especially for classification) is more difficult than human joint detection, which only classify different joints by their relative locations without considering object motion and texture information; 2) The typical FMCW radars on the vehicles have much less resolution than the WiFi array sensors used in RF-Pose. 
As for autonomous or assisted driving applications, Major et al. \cite{major2019vehicle} propose an automotive radar based vehicle detection method using LiDAR information for cross-modal learning. However, our work is different from theirs: 1) They only consider vehicles as the target object class, while we detect pedestrians, cyclists, and cars; 2) The scenarios in their dataset are mostly highways without noisy obstacles, which is easier for radar object detection, while we are dealing with much more diverse driving scenarios. Palffy et al. \cite{palffy2020cnn} propose a radar based, single-frame multi-class object detection method. However, they only consider the data from a single radar frame, which does not involve the object motion information.

\subsection{Datasets}
\label{subsec:related_works_datasets}
Datasets are important to validate the algorithms, especially for the deep learning based methods. Since the first complete autonomous driving dataset, named KITTI \cite{geiger2013vision}, is published, larger and more advanced datasets are now available \cite{apollo_scape_dataset,waymo_open_dataset,nuscenes2019}. However, due to the hardware compatibility and less developed radar perception techniques, most datasets do not incorporate radar signals as a part of their sensor systems. 
Among the available radar datasets, nuScenes \cite{nuscenes2019} and Astyx HiRes2019 \cite{meyer2019automotive} consider radar with good calibration and synchronization with other sensors. But their radar data format is based on sparse radar points that do not contain the useful Doppler and surface texture information of objects. While Oxford Radar RobotCar Dataset \cite{RadarRobotCarDatasetICRA2020} contains dense radar point clouds, it does not provide any object annotations.
After extensive research on the available datasets, we cannot find a suitable one that includes large-scale radar data in RF image format with labeled ground truth. Therefore, we collect our CRUW dataset, which will be introduced in Section~\ref{sec:dataset}.

\section{Radar Object Detection}
\label{sec:rod}

In this section, the student's pipeline of our radar object detection is addressed. First, the raw radar signals are pre-transformed to RF images to be compatible with the image-based convolution neural networks (CNNs). After that, some special properties and challenges of RF images are analyzed. Second, the proposed RODNet is introduced with various functional components. Third, two customized modules are added to the RODNet to handle the above challenges. Finally, a post-processing method, called location-based non-maximum suppression (L-NMS), is adopted to recover ConfMaps for the final detections. 

\subsection{Radar Signal Processing and Properties}
\label{subsec:ramap}

\begin{figure}[t]
\centering
\includegraphics[width=\linewidth]{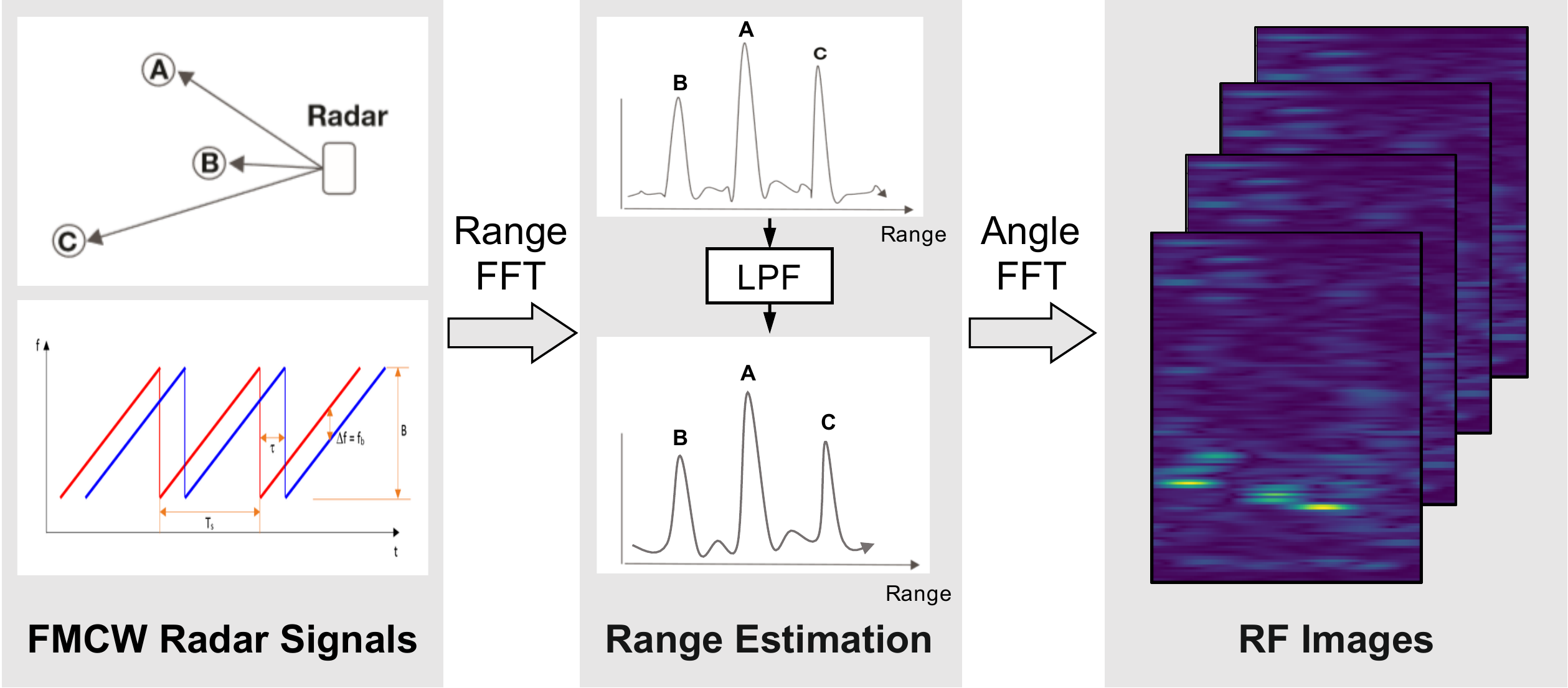}
  \caption{The workflow of the RF image generation from the raw radar signals.}
  \label{fig:ramap_gen}
\end{figure}

In this work, we use a common radar data representation, named radio frequency (RF) image, to represent our radar signal reflections. RF images are in radar range-azimuth coordinates and can be described as a bird's-eye view (BEV) representation, where the $x$-axis denotes azimuth (angle) and the $y$-axis denotes range (distance).
FMCW radar transmits continuous chirps and receives the reflected echoes from the obstacles. After the echoes are received and pre-processed, we implement the fast Fourier transform (FFT) on the samples to estimate the range of the reflections. A low-pass filter (LPF) is then utilized to remove the high-frequency noise across all chirps in each frame at the rate of 30 FPS. After the LPF, we conduct a second FFT on the samples along different receiver antennas to estimate the azimuth angle of the reflections and obtain the final RF images. 
This RF image generation workflow is shown in Fig.~\ref{fig:ramap_gen}. 
After being transformed into RF images, the radar data become a similar format as image sequences, which can thus be directly processed by an image-based CNN.

Moreover, radio frequency data have the following special properties to be handled for the object detection task. 
\begin{itemize}
\item \textbf{Rich motion information.} 
According to the Doppler principle of the radio signal, rich motion information is included. The objects' speed and its law of variation over time is dependent on their surface texture information, size and shape details, etc. 
For example, the motion information of a non-rigid body, like a pedestrian, is usually widely distributed, while for a rigid body, like a car, it should be more consistent due to the Doppler effect. 
To better utilize the temporal information, we need to consider multiple consecutive radar frames, instead of one single frame, as the system input. 
\item \textbf{Inconsistent resolution.} 
Radar usually has high-resolution in range but low-resolution in azimuth angle due to the limitation of radar hardware specifications, like the number of antennas and the distances among them. 
\item \textbf{Complex numbers.} 
Radio signals are usually represented as complex numbers containing frequency and phase information. This kind of data is unusual to be modeled by a typical CNN architecture. 
\end{itemize}

According to the above properties, the proposed radar object detection method needs to have the following capabilities:
1) Extract temporal information; 2) Handle multiple spatial scales; 3) Be able to deal with complex number data. These capabilities.

\begin{figure*}[!t]
\centering
\includegraphics[width=\linewidth]{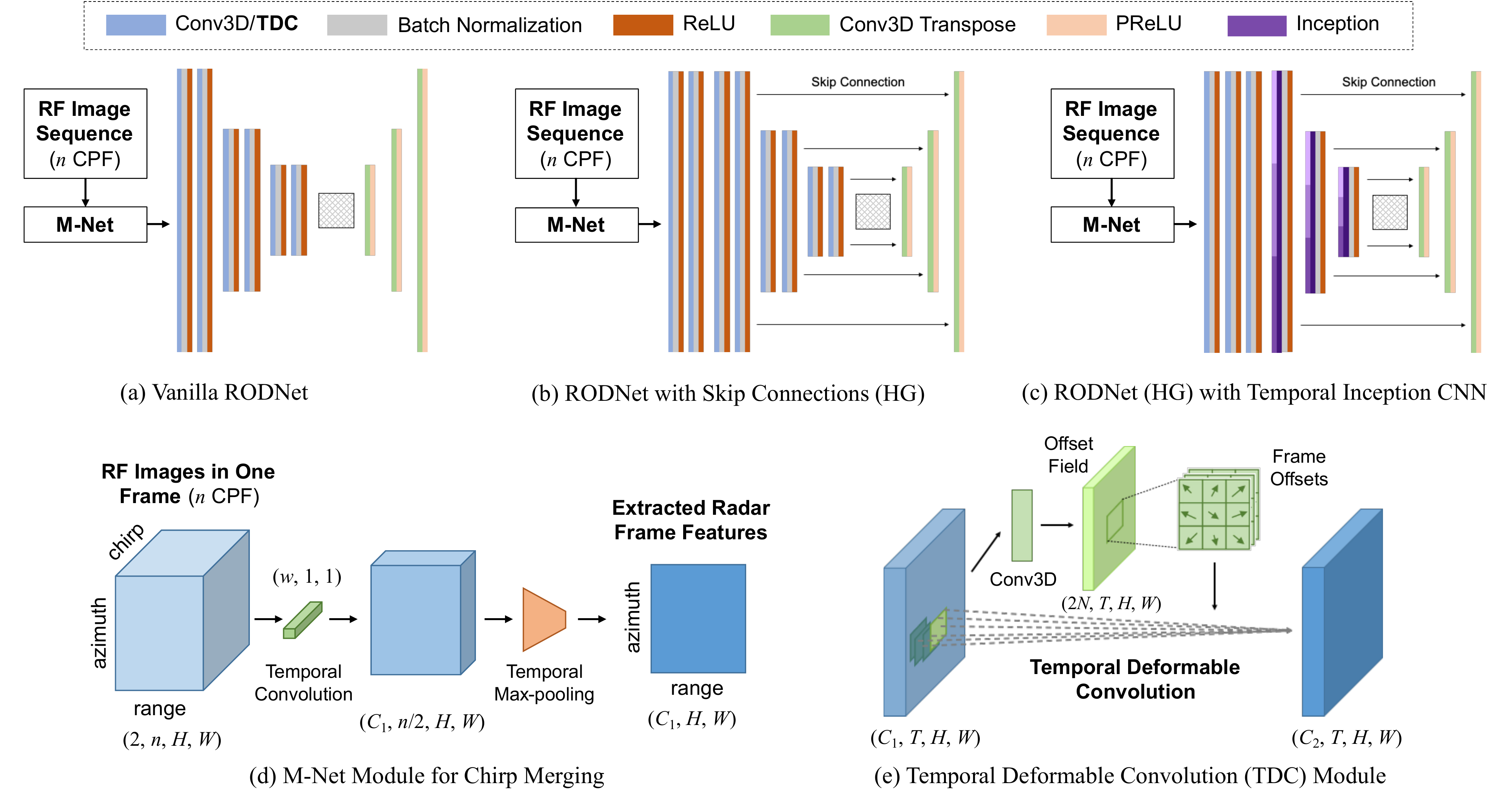}
  \caption{The architecture and modules of our proposed RODNet. Three different components of the RODNet are all implemented based on the 3D CNN/TDC and autoencoder network, as shown in (a), (b), and (c). The input of the RODNet is RF images with $n$ chirps per frame (CPF). When $n=1$, we select only one chirp's data randomly to feed into the RODNet, while for $n>1$, the M-Net module is implemented to merge the data from different chirps in this frame. The M-Net module, described in (d), takes one frame with multiple radar chirps as input and outputs this frame's merged features. Moreover, the temporal deformable convolution (TDC) module in (e) is introduced to handle the radar object dynamic motion within the input RF image sequence.}
  \label{fig:rodnet_arch}
\end{figure*}

\subsection{RODNet Architecture}
\label{subsec:rodnet}

There are three functional components adopted in constructing the network architecture of the RODNet, as shown in Fig.~\ref{fig:rodnet_arch}~(a)-(c), which is implemented based on a 3D CNN with an autoencoder structure. 
More specifically, our RODNet starts with a na\"ive version of a 3D CNN autoencoder network, shown in Fig.~\ref{fig:rodnet_arch}~(a). Then, built upon an hourglass based autoencoder \cite{newell2016stacked}, shown in Fig.~\ref{fig:rodnet_arch}~(b), where skip connections are added to transmit the features directly from bottom layers to top layers.
We further add the temporal inception convolution layers to extract different lengths of temporal features from the input RF image sequence, inspired by the spatial inception convolution layer proposed in \cite{szegedy2015going}, shown in Fig.~\ref{fig:rodnet_arch}~(c). 

The input of our network is a snippet of RF images $\bm{R}$ with dimension $(C_{RF}, T, n, H, W)$, where $C_{RF}$ is the number of channels in each complex-numbered RF images, referring \cite{zhao2018through}, where the real and imaginary values are treated as two different channels in one RF image, i.e., $C_{RF} = 2$; $T$ is the number of RF image frames in the snippet; $n$ is the number of chirps in each frame; $H$ and $W$ are the height and width of the RF images, respectively. 

After passing through the network, ConfMaps $\bm{\hat{D}}$ with dimension $(C_{cls}, T, H, W)$ are predicted, where $C_{cls}$ is the number of object classes. Note that RODNet predicts separate ConfMaps for each object class of radar RF images. With systematically derived binary annotations using the teacher's pipeline described in Section~\ref{sec:cross_modal}, we can train our RODNet using binary cross-entropy loss,
\begin{equation}
    \ell = - \sum_{cls} \sum_{i,j} \bm{D}_{i,j}^{cls} \log \bm{\hat{D}}_{i,j}^{cls} + \left( 1 - \bm{D}_{i,j}^{cls} \right) \log \left( 1 - \bm{\hat{D}}_{i,j}^{cls} \right).
\end{equation}
Here, $\bm{D}$ represents the ConfMaps generated from CRF annotations, $\bm{\hat{D}}$ represents the predicted ConfMaps, $(i,j)$ represents the pixel indices, and $cls$ is the class label.

\subsection{M-Net Module}
\label{subsec:mnet}

Besides the temporal features across different frames in each RF snippet, all the information from different chirps contributes to the features for radar object detection. In order to better integrate this dynamic information from different chirps, we propose a customized module, called M-Net, before the RF snippets are sent into the RODNet. As shown in Fig.~\ref{fig:rodnet_arch}~(d), the RF images of one frame with $n$ chirps are sent into M-Net with a dimension of $(C_{RF}, n, H, W)$, where $C_{RF}=2$. First, a temporal convolution is applied to extract temporal features among the $n$ chirps. This M-Net CNN operation performs like a Doppler compensated FFT to extract dynamic motion features but can be trained end-to-end in the deep learning architecture. Then, to merge the features from $n$ chirps into one, a temporal max-pooling layer is applied. Finally, the output of M-Net is the extracted radar frame features with a dimension of $(C_1, H, W)$, where $C_1$ is the number of filters for the temporal convolution. 
After M-Net is applied to each radar frame, the extracted features from all the frames in the input snippet are concatenated along time and sent as the input to the subsequent RODNet, as shown in Fig.~\ref{fig:rodnet_arch}~(a)-(c).

\subsection{Temporal Deformable Convolution}
\label{subsec:tdc}

As mentioned in Section~\ref{subsec:rodnet}, the input of the RODNet is a snippet of RF images after features are merged by the M-Net. Thus, during this period, locations of the objects in the radar range-azimuth coordinates may be shifted due to object relative motion, which means the reflection patterns in RF images may change their locations within the snippet. 
However, the classical 3D convolution can only capture the static features within a regular cuboid. Therefore it is not the best feature extractor for the RF snippets with object's relative motion. 

Recently, Dai \textit{et al.}~\cite{dai2017deformable} propose a new convolution network, named deformable convolution network (DCN), for image-based object detection to handle the deformed objects within the images. In the deformable convolution, the original convolution grid is deformable in the sense that each grid point is shifted by a learnable offset, and the convolution is operated on these shifted grid points. 

Inspired by DCN, we generalize the deformed kernel to the 3D CNN and name this novel operator as temporal deformable convolution (TDC). 
Use the 3D CNN with kernel size of $(3, 3, 3)$ and dilation $1$ as an example, the regular receptive field $\mathcal{R}$ can be defined as
\begin{equation}
    \mathcal{R} = \{(-1, -1, -1), (-1, 0, 0), \cdots, (0,1,1), (1,1,1)\}.
\end{equation}
For each location $\mathbf{p}_0$ on the output feature map $\mathbf{y}$, the classical 3D convolution can be described as 
\begin{equation}
    \mathbf{y}(\mathbf{p}_0) = \sum_{\mathbf{p}_n \in \mathcal{R}} \mathbf{w}(\mathbf{p}_n) \cdot \mathbf{x}(\mathbf{p}_0 + \mathbf{p}_n),
    \label{eq:classical_cnn}
\end{equation}
where $\mathbf{w}$ is the convolution kernel weight, $\mathbf{x}$ is the input feature map, and $\mathbf{p}_n$ enumerates the locations in $\mathcal{R}$.

In order to handle the object dynamic motion in the temporal domain, we propose TDC by adding an additional offset field $\{\Delta \mathbf{p}_n\}_{n=1}^N$, where $N=|\mathcal{R}|$ is the size of the receptive field. So that Eq.~\ref{eq:classical_cnn} becomes 
\begin{equation}
    \mathbf{y}(\mathbf{p}_0) = \sum_{\mathbf{p}_n \in \mathcal{R}} \mathbf{w}(\mathbf{p}_n) \cdot \mathbf{x}(\mathbf{p}_0 + \mathbf{p}_n + \Delta \mathbf{p}_n).
    \label{eq:tdc}
\end{equation}
Note that the offset field $\Delta \mathbf{p}_n$ is only deformed within each temporal location, i.e., the receptive location of a certain frame will not be deformed to other frames so that the temporal domain of the offset field is always zero. To simplify the implementation process, the offset vectors are defined as 2D vectors so that the overall offset field has a dimension of $(2N, T, H, W)$. An illustration of our proposed TDC is shown in Fig.~\ref{fig:rodnet_arch}~(e).

Similar to \cite{dai2017deformable}, since the offset field $\Delta \mathbf{p}_n$ is typically fractional, Eq.~\ref{eq:tdc} is implemented via bilinear interpolation as 
\begin{equation}
    \mathbf{x}(\mathbf{p}) = \sum_{\mathbf{q}} G(\mathbf{q}, \mathbf{p}) \cdot \mathbf{x}(\mathbf{q}),
\end{equation}
where $\mathbf{p} = \mathbf{p}_0 + \mathbf{p}_n + \Delta \mathbf{p}_n$ is the fractional location; $\mathbf{q}$ enumerates all integer locations in the 3D feature map $\mathbf{x}$; and $G$ is the bilinear interpolation kernel which is also two dimensional in the spatial domain. The back-propagation formulation of TDC is similar to that discussed in \cite{dai2017deformable} except adding the temporal domain, and is described in the supplementary document.

\subsection{Post-processing by Location-based NMS}
\label{subsec:postprocess}

\begin{figure}[t]
    \centering
    \includegraphics[width=.8\linewidth]{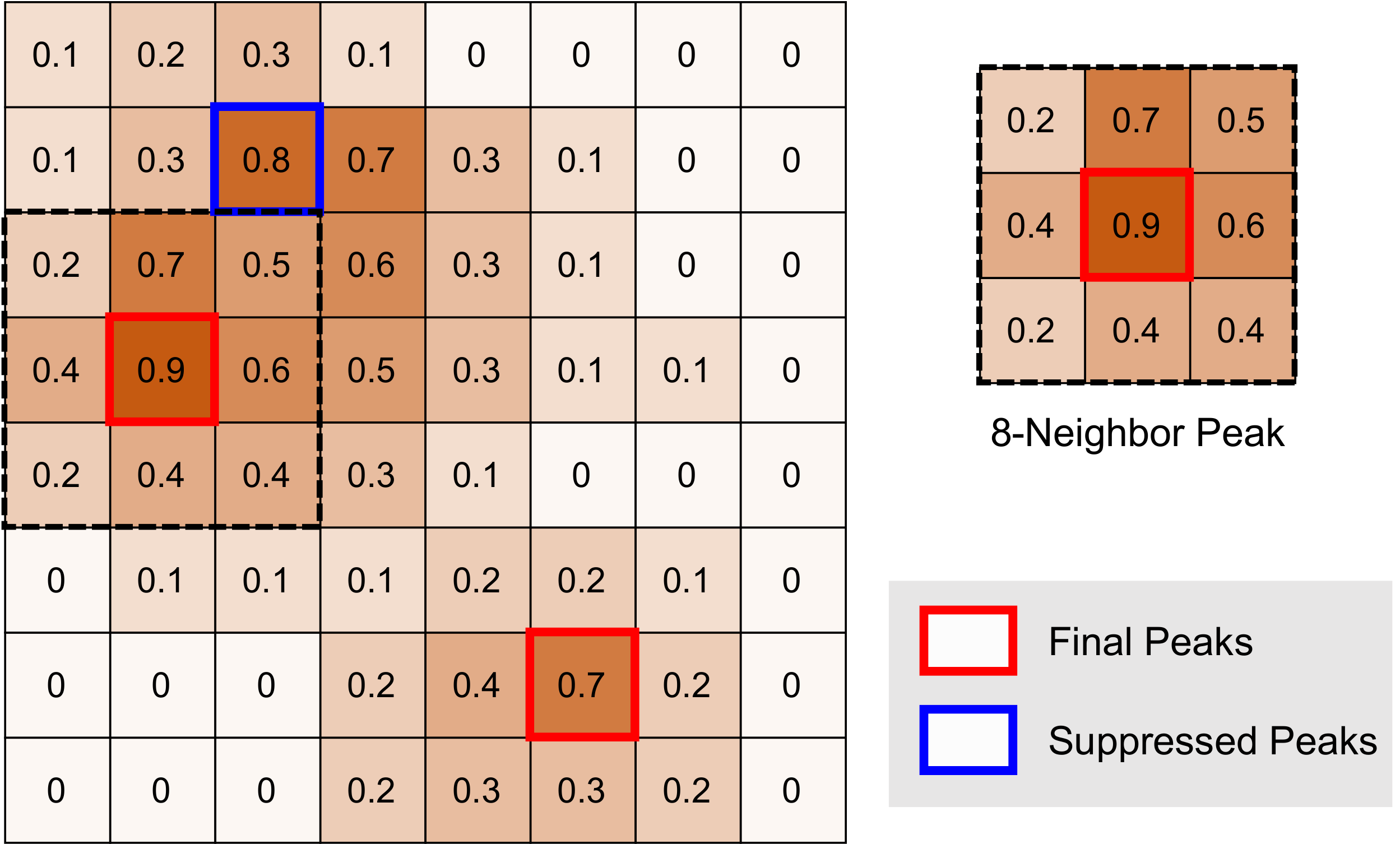}
    \caption{Example for L-NMS on a ConfMap. The numbers represent the confidence scores predicted by the RODNet. The 8-neighbor peaks are first detected, and some peaks are then suppressed if they are nearby some other peaks with higher confidence.}
    \label{fig:lnms}
\end{figure}

After predicting ConfMaps from a given RF snippet, a post-processing step is still required to obtain the final detections. 
Here, we adopt the idea of non-maximum suppression (NMS), which is frequently used in image-based object detection to remove the redundant bounding boxes from the detection results. 
Traditionally, NMS uses intersection over union (IoU) as the criterion to determine if a bounding box should be removed due to its too much overlapping with the detection candidate of the highest confidence. However, there is no bounding box definition in our RF images nor the resulting output ConfMaps. Thus, inspired by object keypoint similarity (OKS) defined for human pose evaluation in the COCO dataset \cite{lin2014microsoft}, we define a new metric, called object location similarity (OLS) that is similar to the role of IoU, to describe the correlation between two detections considering their distance, classes and scale information on ConfMaps. More specifically, 
\begin{equation}
    \text{OLS} = \exp \left\{ \frac{-d^2}{2 (s \kappa_{cls})^2} \right\},
    \label{eq:ols}
\end{equation}
where $d$ is the distance (in meters) between the two points in an RF image; $s$ is the object distance from the radar sensor, representing object scale information; and $\kappa_{cls}$ is a per-class constant that represents the error tolerance for class $cls$, which can be determined by the object average size of the corresponding class. We empirically determine $\kappa_{cls}$ to make OLS distributed reasonably between 0 and 1. 
Here, we try to interpret OLS as a Gaussian distribution, where distance $d$ acts as the bias and $(s \kappa_{cls})^2$ acts as the variance. Therefore, OLS is a metric of similarity, which also considers object sizes and distances, so that more reasonable than other traditional distance metrics, such as Euclidean distance, Mahalanobis distance, etc. 
This OLS metric is also used to \textit{match detections and ground truth} for evaluation purpose, mentioned in Section~\ref{subsec:evaluation}.

After OLS is defined, we propose a location-based NMS (L-NMS) for post-processing. An example of the L-NMS is shown in Fig.~\ref{fig:lnms}, and the procedure can be summarized as follows:
\begin{itemize}
    \item[1)] Get all the 8-neighbor peaks in all $C_{cls}$ channels in ConfMaps within the $3 \times 3$ window as a peak set $P = \{p_n\}_{n=1}^{N}$.
    \item[2)] Pick the peak $p^* \in P$ with the highest confidence, put it to the final peak set $P^*$ and remove it from set $P$. Calculate OLS with each of the rest peaks $p_i$ $(p_i \neq p^*)$.
    \item[3)] If OLS between $p^*$ and $p_i$ is greater than a threshold, remove $p_i$ from the peak set.
    \item[4)] Repeat Steps 2 and 3 until the peak set becomes empty.
\end{itemize}

Moreover, during the inference stage, we can send overlapped RF snippets into the network, which provides different ConfMaps predictions for a single radar frame. Then, we average these different ConfMaps together to obtain the final ConfMaps results. 
This scheme can improve the system's robustness and can be considered as a performance-speed trade-off, which will be further discussed in Section~\ref{subsec:ablation}.

\section{Cross-Modal Supervision}
\label{sec:cross_modal}

\begin{figure}[t]
    \centering
    \includegraphics[width=.95\linewidth]{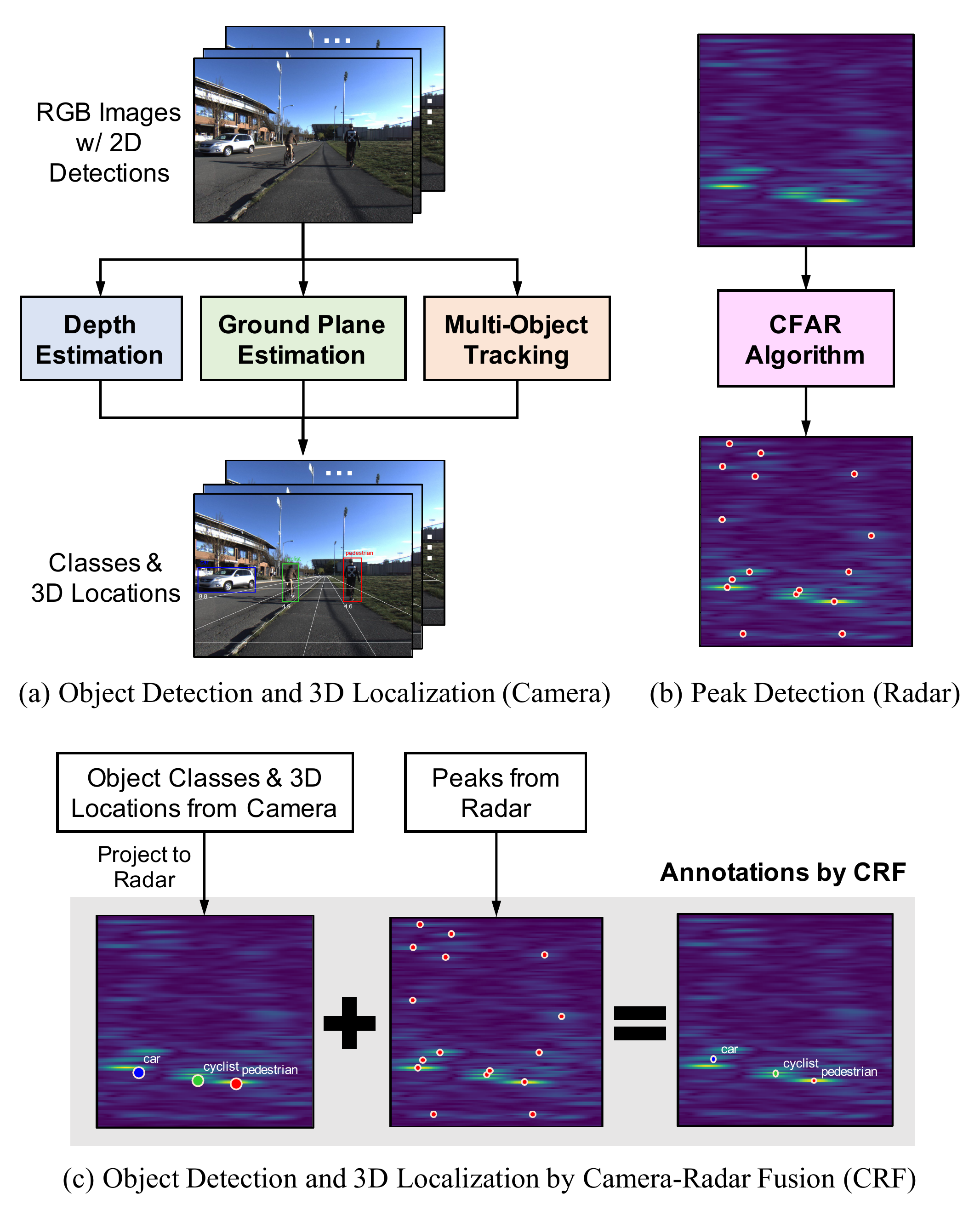}
    \caption{Three teacher's pipelines for cross-model supervision. (a) Camera-only method that provides object classes and 3D locations; (b) Radar-only method that only provides peak locations without object class; (c) Camera-radar fusion method that provides object classes and more accurate 3D locations.}
    \label{fig:cross_sup}
\end{figure}

In this section, the teacher's pipeline that provides supervision for the RODNet is described. First, the annotation methods by camera-only and camera-radar fusion are introduced to obtain the accurate object classes and 3D locations. Then, these annotations are applied to ConfMaps as ground truth for training the network end-to-end.

\subsection{Camera-Only (CO) Supervision}
\label{subsec:cam_anno}

Object detection and 3D localization have been explored by researchers in computer vision community for many years. Here, we use Mask R-CNN \cite{he2017mask} as our image-based object detector, which can provide object classes, bounding boxes, and their instance masks. 
While, object 3D localization is more challenging due to the loss of the depth information during the camera projection from the 3D world into 2D images. To recover the 3D information from 2D images, we take advantage of a recent work on an effective and robust system for visual object 3D localization based on a monocular camera \cite{wang2019monocular}. 
The proposed system takes a CNN inferred depth map as the input, incorporating adaptive ground plane estimation and multi-object tracking results, to effectively estimate object classes and 3D locations relative to the camera. 
The main strength of this camera-only system is that it can robustly estimate objects' 3D locations purely from monocular videos, resulting in a very low requirement on the visual data used for annotation, i.e., one single camera is good enough to obtain the annotation results. 
The rational behind this capability can be described in three folds: 1) Depth estimation used in the system is self-supervised by the stereo image pairs during training, so as to provide the absolute scale information missing in the monocular camera systems; 2) Adaptive ground plane estimation, based on sparse detected object feet points and dense semantic segmented ground points, is proposed to handle inaccurate depth within each frame; 3) A multi-object tracking technique is incorporated to address the inter-frame bias issue and temporally smooth the object 3D trajectories. 
A simplified illustration of the monocular camera object 3D localization system proposed in \cite{wang2019monocular} is shown in Fig.~\ref{fig:cross_sup}~(a).
Stereo cameras can also be used for object 3D localization, however, high computational cost and sensitivity to camera setup configurations (e.g., baseline distance) result in the limitation of the stereo object 3D localization system. 

However, as also observed in \cite{wang2020rodnet}, the above camera-only system may not be accurate enough after transforming to the radar's range-azimuth coordinates because: 1) The systematic bias in the camera-radar sensor system that the peaks in the RF images may not be consistent with the 3D geometric center of the object; 2) Cameras' performance can be easily affected by lighting or weather conditions. Since we do have the radar information available, camera-radar cross-calibration and supervision should be used. Therefore, an even more accurate self-annotation method, based on camera-radar fusion, is required for training the RODNet.

\subsection{Camera-Radar Fusion (CRF) Supervision}
\label{subsec:crf_anno}

An intuitive way of improving the above camera-only annotation is by taking advantage of radar, which has a plausible capability of range estimation without any systematic bias. Here, we adopt the Constant False Alarm Rate (CFAR) detection algorithm \cite{richards2005fundamentals}, which is commonly used in signal processing to detect peaks in the RF image. As shown in Fig.~\ref{fig:cross_sup}~(b), CFAR algorithm can detect several peaks in the RF image, denoted as red dots. However, these detected peaks cannot be directly used as supervision because 1) CFAR algorithm cannot provide the object classes for each detection; 2) CFAR algorithm usually gives a large number of false positive detections. Thus, an object localization method by camera-radar fusion strategy is needed to address these issues. 

Fig.~\ref{fig:cross_sup}~(c) illustrates the camera-radar fusion (CRF) pipeline, where the classes and 3D locations of the detected objects from the camera are first passed through a transformation to project the detections from 3D camera coordinates to radar range-azimuth coordinates. The transformation can be formulated as 
\begin{equation}
    \begin{aligned}
    &\rho_c = \sqrt{(x^{c}-x_{or})^2 + (z^{c}-z_{or})^2}, \\
    &\theta_c = \tan^{-1} \left(\frac{x^{c}-x_{or}}{z^{c}-z_{or}}\right),
    \end{aligned}
\end{equation}
where $(\rho_c, \theta_c)$ denotes the projected location in radar range-azimuth coordinates; $(x^c, z^c)$ denotes the object location in the camera BEV coordinates; and $(x_{or}, z_{or})$ denotes the location of radar origin in the camera BEV coordinates, aligned from the sensor system calibration.
The peak detections from the CFAR algorithm are also involved in the same radar range-azimuth coordinates. Finally, the fusion algorithm is applied to estimate the final annotations on the input RF image. 

After the coordinates between the camera and radar are aligned, a probabilistic CRF algorithm is further developed to achieve a more reliable and systematic annotation performance. 
The basic idea of this algorithm is to generate two probability maps for camera and radar locations separately, and then fuse them by element-wise product. The probability map for camera locations with object class $cls$ is generated by
\begin{equation*}
\resizebox{0.49\textwidth}{!}{$
    \mathcal{P}^c_{(cls)}(\mathbf{x}) = \max_{i} \left\{ \mathcal{N} \left( \frac{1}{2 \pi \sqrt{|\mathbf{\Sigma}^c_{i(cls)}|}} \exp \left\{ -\frac{1}{2} (\mathbf{x} - \mathbf{\mu}^c_{i})^{\top} (\mathbf{\Sigma}^c_{i(cls)})^{-1} (\mathbf{x} - \mathbf{\mu}^c_{i}) \right\} \right) \right\},
$}
\end{equation*}
\begin{equation}
\resizebox{0.32\textwidth}{!}{$
    \mathbf{\mu}^c_{i} = 
    \begin{bmatrix}
    \rho^c_{i} \\
    \theta^c_{i} 
    \end{bmatrix}, 
    \mathbf{\Sigma}^c_{i(cls)} = 
    \begin{bmatrix}
    \left(d_i s_{(cls)} / c_i \right)^2 & 0 \\
    0 & \delta_{(cls)} 
    \end{bmatrix}.
$}
\end{equation}
Here, $d_i$ is the object depth, $s_{(cls)}$ is the scale constant, $c_i$ is the depth confidence, and $\delta_{(cls)}$ is the typical azimuth error for camera localization. $\mathcal{N}(\cdot)$ represents the normalization operation for each object's probability map. Similarly, the probability map for radar locations is generated by
\begin{equation*}
\resizebox{0.46\textwidth}{!}{$
    \mathcal{P}^r(\mathbf{x}) = \max_{j} \left\{ \mathcal{N} \left( \frac{1}{2 \pi \sqrt{|\mathbf{\Sigma}^r_{j}|}} \exp \left\{ -\frac{1}{2} (\mathbf{x} - \mathbf{\mu}^r_{j})^{\top} (\mathbf{\Sigma}^r_{j})^{-1} (\mathbf{x} - \mathbf{\mu}^r_{j}) \right\} \right) \right\},
$}
\end{equation*}
\begin{equation}
\resizebox{0.22\textwidth}{!}{$
    \mathbf{\mu}^r_{j} = 
    \begin{bmatrix}
    \rho^r_{j} \\
    \theta^r_{j} 
    \end{bmatrix}, 
    \mathbf{\Sigma}^r_{j} = 
    \begin{bmatrix}
    \delta^r_{j} & 0 \\
    0 & \epsilon(\theta^r_{j}) 
    \end{bmatrix}.
$}
\end{equation}
Here, $\delta^r_j$ is the radar's range resolution, and $\epsilon(\cdot)$ is the radar's azimuth resolution. 
Then, an element-wise product is used to obtain the fused probability map for each class,
\begin{equation}
    P^{CRF}_{(cls)}(\mathbf{x}) = P^c_{(cls)}(\mathbf{x}) * P^r(\mathbf{x}).
\end{equation}
Finally, the fused annotations are derived from the fused probability maps $P^{CRF}$ by peak detection.

The object 3D localization accuracy of both CO and CRF annotation is discussed later in Section~\ref{subsec:results}.

\subsection{ConfMap Generation}
\label{subsec:confmap}

After objects are accurately localized in the radar range-azimuth coordinates, we need to transform the results into a proper representation that is compatible with our RODNet. Considering the idea in \cite{cao2017realtime} that defines the human joint heatmap to represent joint locations, we define the confidence map (ConfMap) in range-azimuth coordinates to represent object locations. One set of ConfMaps has multiple channels, where each channel represents one specific class label, i.e., car, pedestrian, and cyclist. The value at the pixel in the $cls$-th channel represents the probability of an object with class $cls$ occurring at that range-azimuth location. 
Here, we use Gaussian distributions to set the ConfMap values around the object locations, 
whose mean is the object location, and the variance is related to the object class and scale information.

\section{CRUW Dataset}
\label{sec:dataset}

\begin{table}[t]
    \centering
    \caption{Sensor Configurations for CRUW Dataset.}
    \begin{tabular}{l c | l c}
        \hline
        Camera & Value & Radar & Value \\
        \hline
        Frame rate & 30 FPS & Frame rate & 30 FPS \\
        Pixels (H$\times$W) & 1440$\times$1080 & Frequency & 77 GHz \\
        Resolution & 1.6 MegaPixels & \# of transmitters & 2 \\
        Field of View & 93.6$^{\circ}$ & \# of receivers & 4 \\
        Stereo Baseline & 0.35 m & \# of chirps per frame & 255 \\
        && Range resolution & 0.23 m \\
        && Azimuth resolution & $\sim$15$^{\circ}$ \\
        \hline
    \end{tabular}
    \label{tab:sensor_config}
\end{table}

\begin{figure}[!t]
\centering
\includegraphics[width=\linewidth]{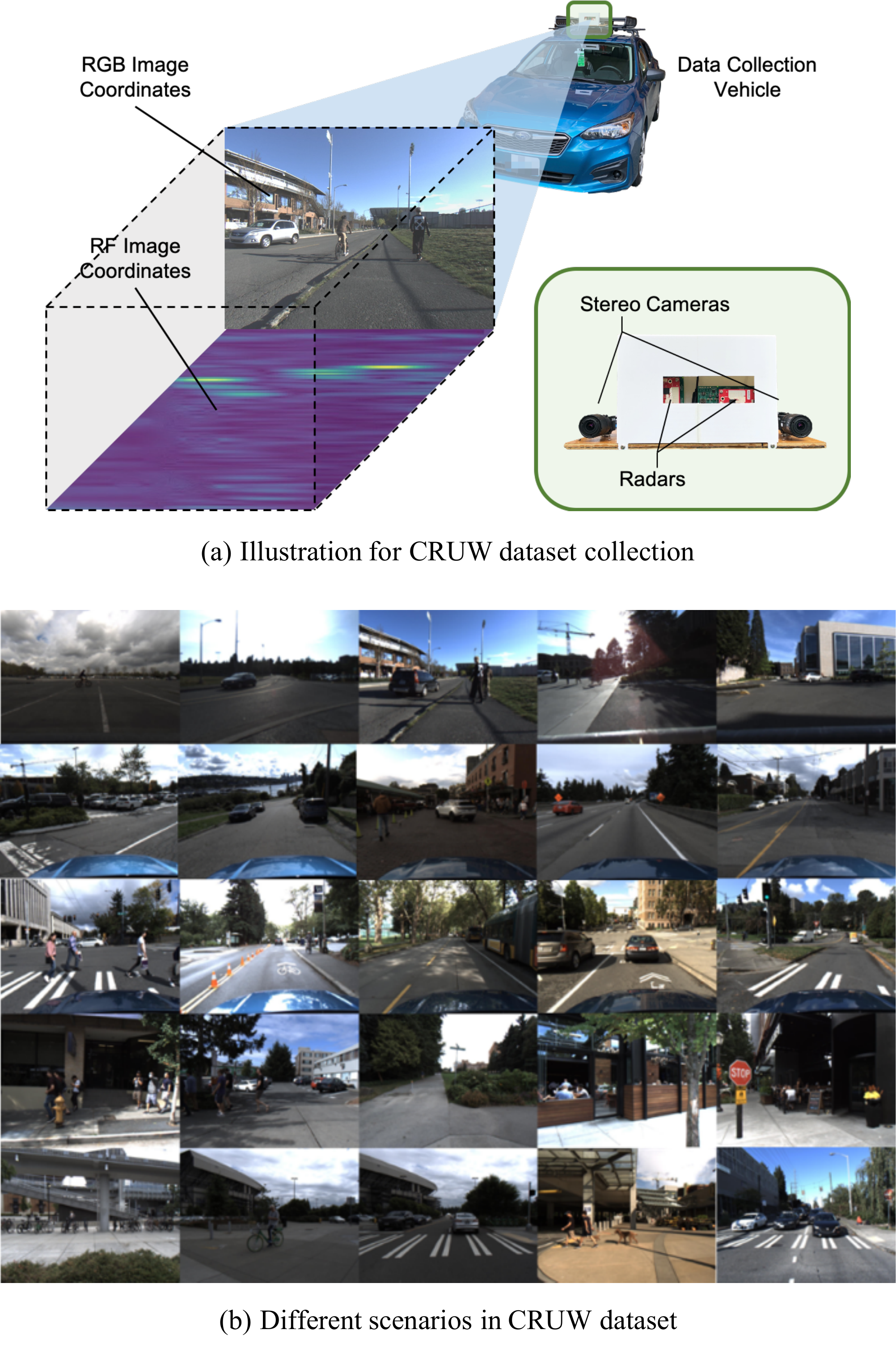}
  \caption{The sensor platform and driving scenarios for our CRUW dataset.
  }
  \label{fig:dataset}
\end{figure}

Going through some existing datasets for autonomous driving discussed in Section~\ref{subsec:related_works_datasets}, the 3D radar point format is commonly used. While, it does not contain the discriminating object motion and surface texture information that is needed for our radar object detection task.  
To efficiently train and evaluate our RODNet using radar data, we collect a new dataset, named Camera-Radar of the University of Washington (CRUW), which uses the format of RF images for the radar data, as mentioned in Section~\ref{subsec:ramap}.
Our sensor platform contains a pair of stereo cameras \cite{flir} and two perpendicular 77GHz FMCW MMW radar antenna arrays \cite{ti}. The sensors, assembled and mounted together as shown in Fig.~\ref{fig:dataset}~(a), are well-calibrated and synchronized. Some configurations of our sensor platform in shown in Table~\ref{tab:sensor_config}. Even though our final cross-modal supervision requires just one monocular camera, the stereo cameras are setup to provide ground truth of depth for performance validation of the CRF supervision.

\begin{figure}[t]
\centering
  \includegraphics[width=\linewidth]{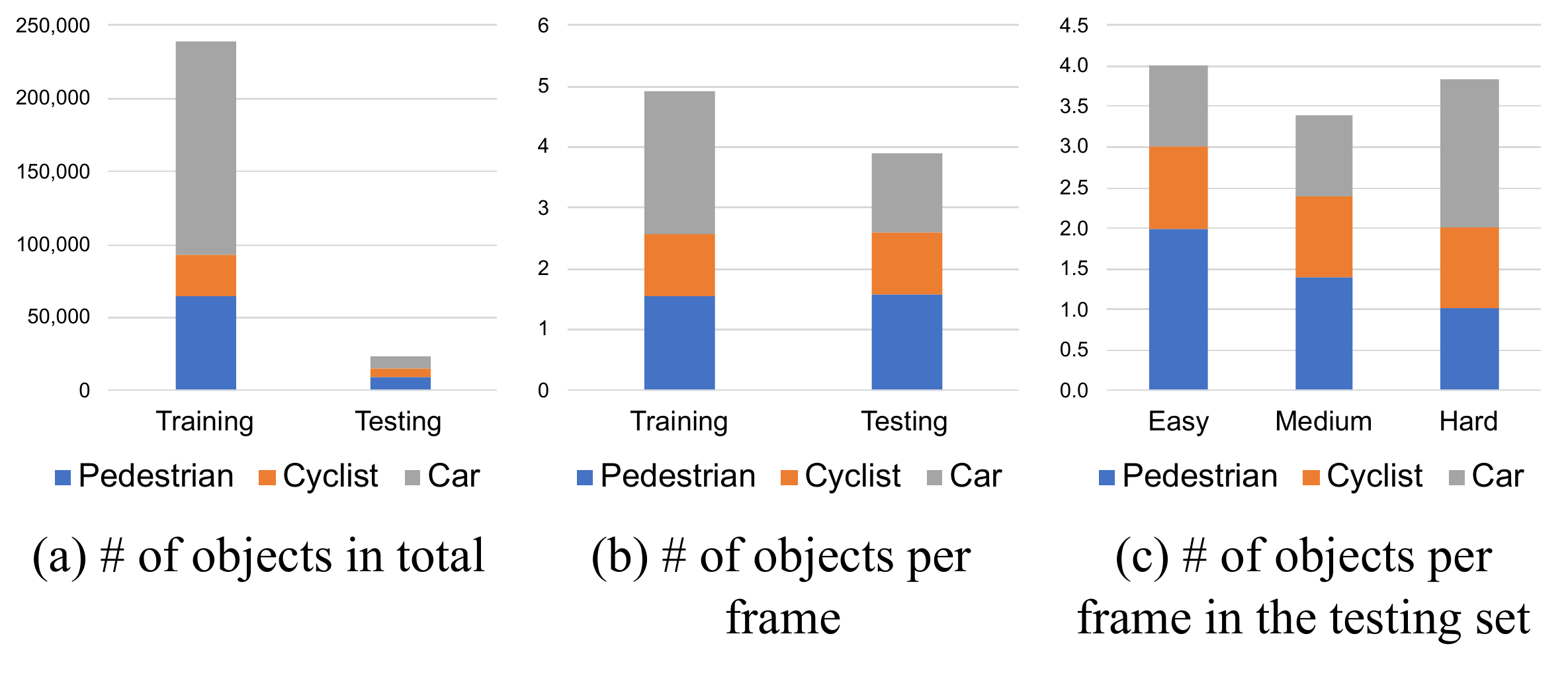}
   \caption{Illustration for our CRUW dataset distribution. Here, (a)-(c) show the object distribution in the radar's FoV ($0$-$25$m, $\pm90^{\circ}$); (d) shows the distribution under different driving scenarios and lighting conditions.}
\label{fig:cr_distribution}
\end{figure}

\begin{table}[t]
    \centering
    \caption{Driving scenarios statistics for CRUW dataset.}
    \begin{tabular}{c c c c}
        \hline
        Scenarios & \# of Seqs & \# of Frames & Vision-Hard \% \\
        \hline
        Parking Lot & 124 & 106K & 15\% \\
        Campus Road & 112 & 94K & 11\% \\
        City Street & 216 & 175K & 6\% \\
        Highway & 12 & 20K & 0\% \\
        \hline
        Overall & 464 & 396K & 9\% \\
        \hline
    \end{tabular}
    \label{tab:scenarios_distri}
\end{table}

\begin{figure*}[!t]
    \centering
    \includegraphics[width=\linewidth]{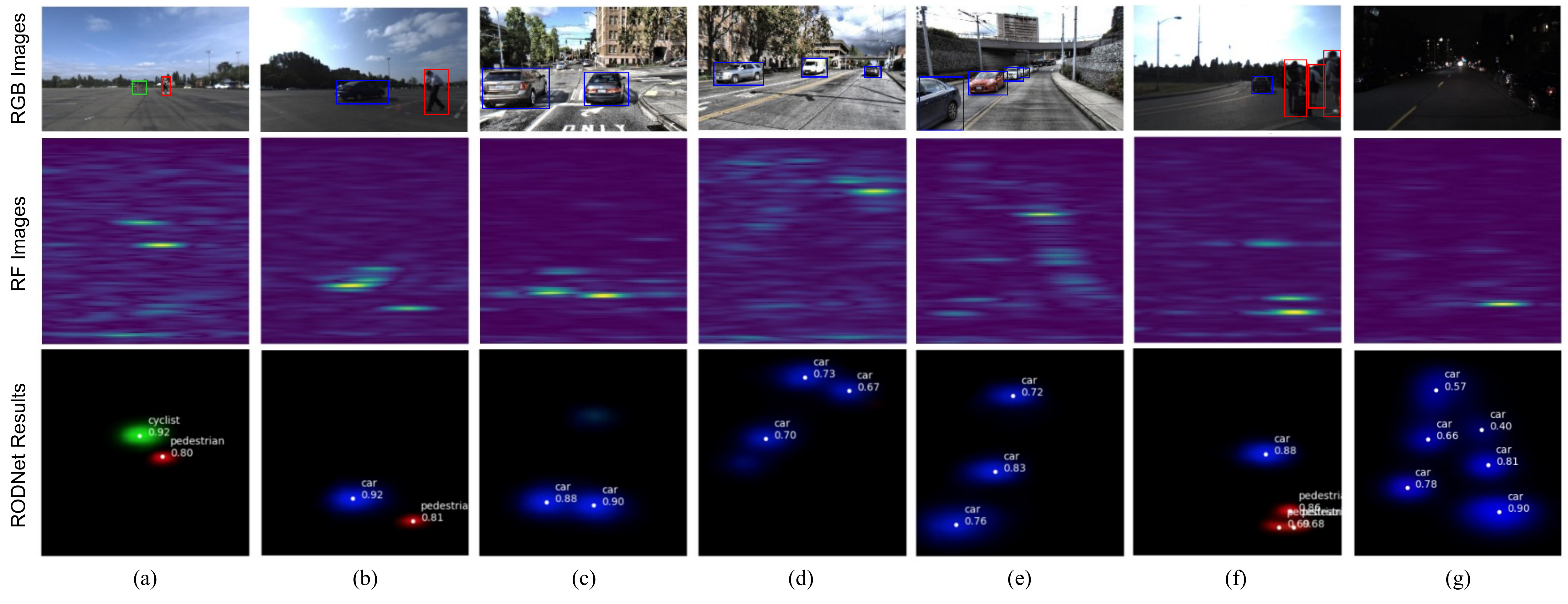}
    \caption{Examples of detection results from our RODNet. The first row shows the RGB images, and the second row shows the corresponding RF images. The ConfMaps predicted by the RODNet are shown in the third row, where the white dots represent the final detections after post-processing. Different colors represent different detected object classes (red: pedestrian; green: cyclist; blue: car). Various driving scenarios are shown, i.e., clear parking lot, crowded city street, and strong/weak lighting condition. More qualitative results are presented in the supplementary materials.}
\label{fig:res_qual_vis}
\end{figure*}

The CRUW dataset contains 3.5 hours with 30 FPS (about 400K frames) of camera-radar data in different driving scenarios, including campus road, city street, highway, and parking lot. Some sample scenarios are shown in Fig.~\ref{fig:dataset}~(b). The data are collected in two different views, i.e., driver front view and driver side view, to ensure that our method is applicable to different perspective views for autonomous or assisted driving. Besides, we also collect several vision-hard sequences of poor image quality, i.e., weak/strong lighting, blur, etc. These data are only used for testing to illustrate that our method can still be reliable when vision techniques most likely fail. 

The data distribution of CRUW is shown in Fig.~\ref{fig:cr_distribution}. The object statistics in (a)-(c) only consider the objects within the radar's field of view (FoV), i.e., $0$-$25$m, $\pm 90^{\circ}$, based on the current hardware capability. There are about 260K objects in CRUW dataset in total, including $92\%$ for training and $8\%$ for testing. The average number of objects in each frame is similar between training and testing data. As for the testing set, we split it into three difficulty levels, i.e., easy, medium, and hard, to evaluate the performance under different scenarios. The criteria for this split include the number of objects, clear/noisy background, normal/extreme lighting, and object relative motion. The four different driving scenarios included in the CRUW dataset are shown in (d) with the number of sequences, frames, and vision-hard percentages. 
From each scenario, we randomly select several complete sequences as testing sequences, which are not used for training. Thus, the training and testing sequences are captured at different locations and at different times.
For the ground truth needed for evaluation purposes, we annotate $10\%$ of the visible and $100\%$ of the vision-hard data. The annotations are operated on the RF images by labeling the object classes and locations according to the corresponding RGB and RF images.

\section{Experiments}
\label{sec:experiments}

\subsection{Evaluation Metrics}
\label{subsec:evaluation}

To evaluate the performance, we utilize our proposed object location similarity (OLS) (see Eq.~\ref{eq:ols}) in Section~\ref{subsec:postprocess}, replacing the role of IoU widely used in image-based object detection, to determine how well a detection result can be matched with a ground truth.
During the evaluation, we first calculate OLS between each detection result and ground truth in every frame. Then, we use different thresholds from $0.5$ to $0.9$ with a step of $0.05$ for OLS and calculate the average precision (AP) and average recall (AR) for different OLS thresholds, which represent different localization error tolerance for the detection results. Here, we use AP and AR to represent the average values among all different OLS thresholds from $0.5$ to $0.9$ and use $\text{AP}^{\text{OLS}}$ and $\text{AR}^{\text{OLS}}$ to represent the values at a certain OLS threshold. Overall, we use AP and AR as our main evaluation metrics for the radar object detection task.

\subsection{Radar Object Detection Results}
\label{subsec:results}

\begin{table*}[t]
\centering
\caption{Radar object detection performance evaluated on CRUW dataset.}
\begin{tabular}{L{2.6cm}|C{0.8cm}C{0.8cm}|C{0.8cm}C{0.8cm}|C{0.8cm}C{0.8cm}|C{0.8cm}C{0.8cm}}
\hline
\multirow{2}*{Methods} & \multicolumn{2}{c|}{Overall} & \multicolumn{2}{c|}{Easy} & \multicolumn{2}{c|}{Medium} & \multicolumn{2}{c}{Hard} \\
\cline{2-9}
~ & AP & AR & AP & AR & AP & AR & AP & AR  \\
\hline
\specialrule{0em}{1pt}{1pt}
Decision Tree \cite{gao2019experiments} & $4.70$ & $44.26$ & $6.21$ & $47.81$ & $4.63$ & $43.92$ & $3.21$ & $37.02$ \\
CFAR+ResNet \cite{8468324} &  $40.49$ & $60.56$ & $78.92$ & $85.26$ & $11.00$ & $33.02$ & $6.84$ & $36.65$\\
CFAR+VGG-16 \cite{gao2019experiments} &  $40.73$ & $72.88$ & $85.24$ & $88.97$ & $47.21$ & $62.09$ & $10.97$ & $45.03$ \\
\specialrule{0em}{1pt}{1pt}
\hline
\specialrule{0em}{1pt}{1pt}
\textbf{RODNet (Ours)} & $\mathbf{85.98}$ & $\mathbf{87.86}$ & $\mathbf{96.97}$ & $\mathbf{98.02}$ & $\mathbf{76.11}$ & $\mathbf{78.57}$ & $\mathbf{67.28}$ & $\mathbf{72.60}$ \\
\specialrule{0em}{1pt}{1pt}
\hline
\end{tabular}
\label{tab:results_main}
\end{table*}

\begin{table*}[t] 
\centering
\caption{Ablation studies on the performance improvement by several functional components in the RODNet.}
\begin{tabular}{L{1.1cm}|C{1.3cm}C{0.8cm}C{0.6cm}C{1cm}||C{0.8cm}|C{0.8cm}C{0.8cm}C{0.8cm}||C{0.8cm}|C{0.8cm}C{0.8cm}C{0.8cm}}
\hline
\specialrule{0em}{1pt}{1pt}
Backbone & Supervision & M-Net & TDC & Inception & AP & $\text{AP}^{0.5}$ & $\text{AP}^{0.7}$ & $\text{AP}^{0.9}$ & AR & $\text{AR}^{0.5}$ & $\text{AR}^{0.7}$ & $\text{AR}^{0.9}$\\
\specialrule{0em}{1pt}{1pt}
\hline
\specialrule{0em}{1pt}{1pt}
\multirow{4}{*}{Vanilla} & CO & & & & $52.62$ & $78.21$ & $54.66$ & $18.92$ & $63.95$ & $84.13$ & $68.76$ & $30.71$  \\
& CRF & & & & $74.29$ & $78.42$ & $76.06$ & $64.58$ & $77.85$ & $80.05$ & $78.93$ & $71.72$ \\
& CRF & \checkmark & & & $78.36$ & $82.73$ & $81.03$ & $65.82$ & $81.54$ & $84.51$ & $83.39$ & $73.53$ \\
& CRF & \checkmark & \checkmark & & $79.86$ & $84.08$ & $82.37$ & $66.74$ & $82.85$ & $86.06$ & $84.43$ & $73.93$  \\
\specialrule{0em}{1pt}{1pt}
\hline
\specialrule{0em}{1pt}{1pt}
\multirow{7}{*}{Hourglass} & CO & & & & $73.86$ & $80.34$ & $74.94$ & $61.16$ & $79.87$ & $83.94$ & $80.73$ & $71.39$  \\
& CO & & & \checkmark & $77.75$ & $82.88$ & $79.93$ & $61.88$ & $81.11$ & $85.13$ & $82.78$ & $68.63$  \\
& CRF & & & & $81.10$ & $84.71$ & $83.08$ & $70.21$ & $84.26$ & $86.54$ & $85.42$ & $77.44$  \\
& CRF & \checkmark & & & $83.37$ & $87.51$ & $\underline{86.04}$ & $71.11$ & $85.64$ & $88.55$ & $87.19$ & $77.37$  \\
& CRF & & & \checkmark & $83.76$ & $87.99$ & $86.00$ & $70.88$ & $85.62$ & $88.79$ & $87.37$ & $76.26$  \\
& CRF & \checkmark & \checkmark & & $\underline{84.38}$ & $\underline{88.69}$ & $85.73$ & $\underline{73.31}$ & $\underline{86.97}$ & $\underline{89.67}$ & $\underline{88.14}$ & $\underline{79.59}$  \\
& CRF & \checkmark & \checkmark & \checkmark & $\mathbf{85.98}$ & $\mathbf{88.77}$ & $\mathbf{87.78}$ & $\mathbf{76.34}$ & $\mathbf{87.86}$ & $\mathbf{89.93}$ & $\mathbf{89.02}$ & $\mathbf{81.26}$  \\
\specialrule{0em}{1pt}{1pt}
\hline
\end{tabular}
\label{tab:results_apolss}
\end{table*}

We train our RODNet using the training data with CRF annotations in the CRUW dataset. For testing, we perform inference and evaluation on the human-annotated data. 
The quantitative results are shown in Table~\ref{tab:results_main}. 
We compare our RODNet results with the following baselines that also use radar-only inputs: 1) A decision tree using some handcrafted features from radar data \cite{gao2019experiments}; 2) CFAR detection is first implemented, and a radar object classification network with ResNet backbone \cite{8468324} is appended; 3) Similar with 2), a radar object classification network with VGG-16 backbone based on CFAR detections mentioned in \cite{gao2019experiments}.
Among all the three competing methods, the AR performance for \cite{gao2019experiments,8468324} is relatively stable in all three different test sets, but their APs vary a lot. Especially, the APs drop from around $80\%$ to $10\%$ for easy to hard testing sets. This is caused by a large number of false positives detected by the traditional CFAR algorithm, which would significantly decrease the precision. 

Comparing with the above baseline methods, our RODNet outperforms significantly on both AP and AR metrics, achieving the best performance of $85.98\%$ AP and $87.86\%$ AR, especially the sustained performance on the medium and hard testing sets, which shows the robustness to noisy scenarios. Note that the results of RODNet shown in Table~\ref{tab:results_main} include all the components proposed for RODNet, i.e., CRF supervision, M-Net, TDC, and temporal inception CNN.

Some qualitative results are shown in Fig.~\ref{fig:res_qual_vis}, where we can find that the RODNet can accurately localize and classify multiple objects in different scenarios. The examples in Fig.~\ref{fig:res_qual_vis} consist of RGB and RF image pairs as well as RODNet detection results under different driving scenarios and conditions, including parking lot, campus road, and city street, with different lighting conditions. Some other examples to show the special strengths of our RODNet are shown in Section~\ref{subsec:strength}.

\begin{figure}[t]
\centering
\includegraphics[width=0.82\linewidth]{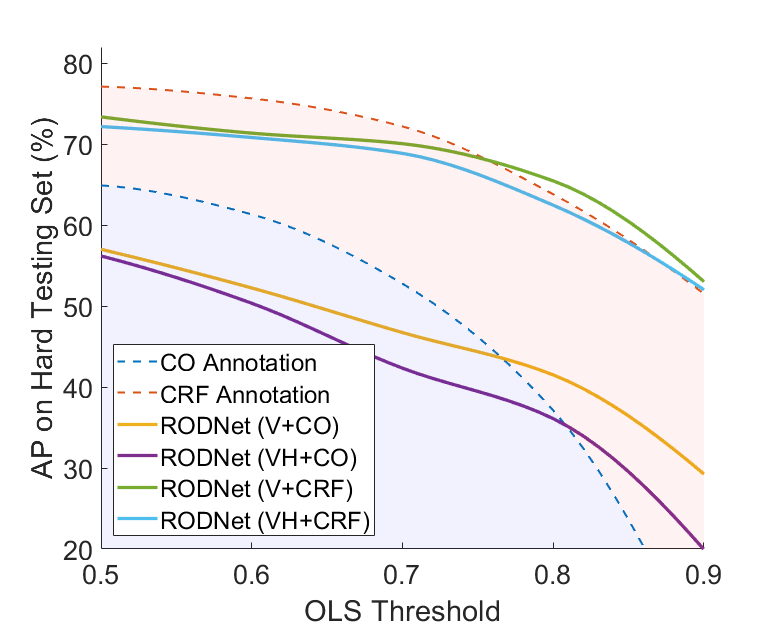}
  \caption{Performance of vision-based and our RODNet on hard testing set with different OLS thresholds, representing localization error tolerance. (CO: camera-only; CRF: camera-radar fusion; V: visible data; VH: vision-hard data.)}
  \label{fig:results_vrcomp}
\end{figure}

To illustrate our teacher's pipeline is qualified for this cross-supervision task, we evaluate the object 3D localization performance for both CO and CRF annotations in Table~\ref{tab:res_obj3dloc}. 
Besides, we also compare the performance between CO/CRF supervision and our RODNet on both visible (V) and vision-hard (VH) data. The results are shown in Fig.~\ref{fig:results_vrcomp} with respect to different OLS thresholds. 
From Fig.~\ref{fig:results_vrcomp}, the performance of the vision-based method drops significantly given a tighter OLS threshold, while our RODNet shows its superiority and robustness on its localization performance. 
Moreover, the RODNet can still maintain the performance on vision-fail data where the vision-based methods have a hard time to maintain the performance.

\begin{table}[t]
    \centering
    \caption{The mean localization error (standard deviation) of CO/CRF annotations on CRUW dataset (in meters).}
    \begin{tabular}{c C{1.6cm} C{1.6cm} C{1.6cm}}
        \hline
        Supervision & Pedestrian & Cyclist & Car \\
        \hline
        CO & $0.69\ (\pm 0.77)$ & $0.87\ (\pm 0.89)$ & $1.57\ (\pm 1.12)$ \\
        CRF & $0.67\ (\pm 0.55)$ & $0.82\ (\pm 0.59)$ & $1.26\ (\pm 0.64)$ \\
        \hline
    \end{tabular}
    \label{tab:res_obj3dloc}
\end{table}

\subsection{Ablation Studies}
\label{subsec:ablation}

In this section, we analyze the performance-speed trade-off of several components of the RODNet and dive into some details to show how our method can accomplish this radar object detection task very well.

\subsubsection{Performance-Speed Trade-off of the RODNet}

\begin{figure}[t]
\centering
\includegraphics[width=0.85\linewidth]{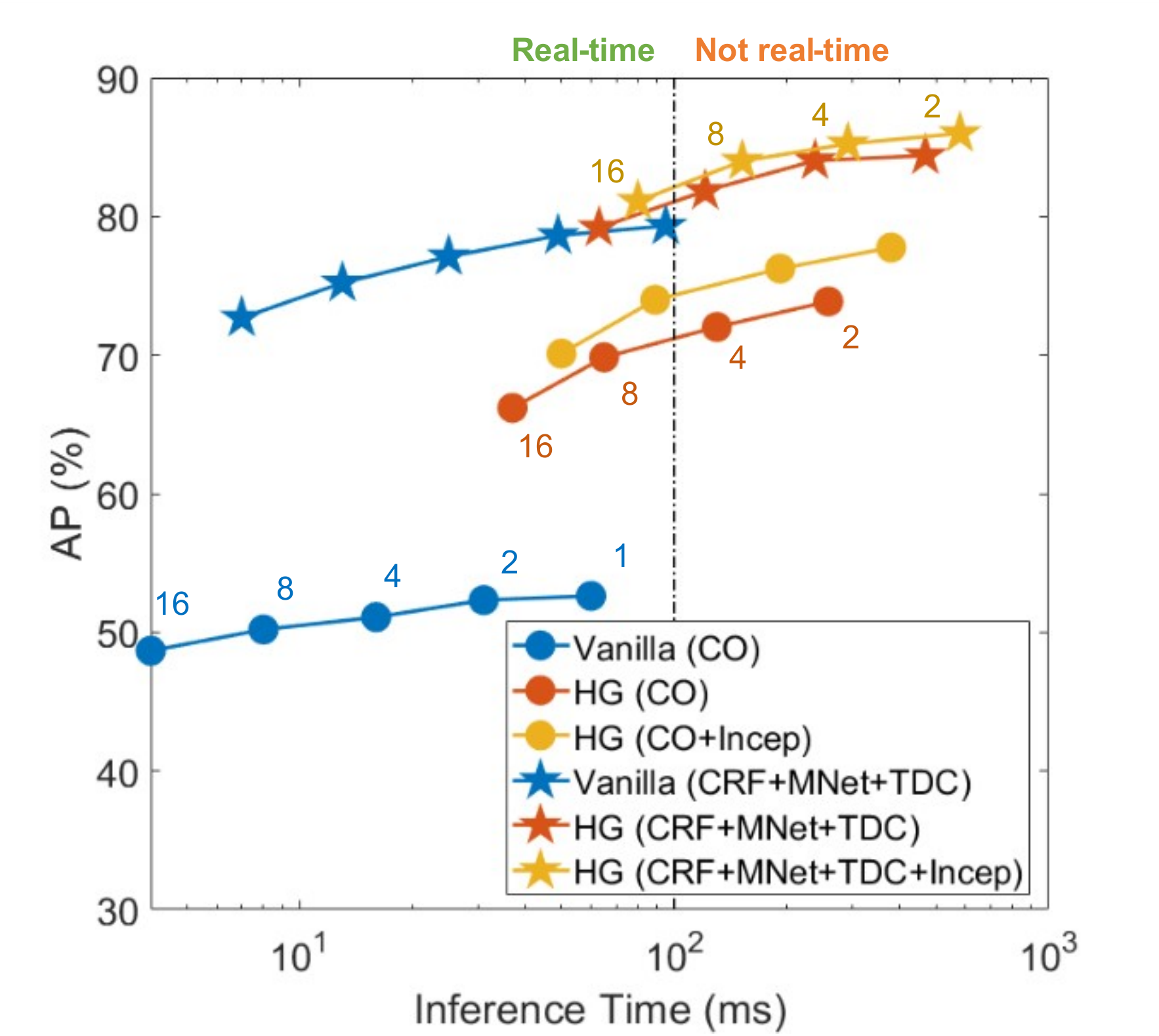}
  \caption{Performance-speed trade-off for real-time implementation. Here, the inference time of less than 100 ms is used as the real-time criterion. We use the snippet length of $16$, and the numbers beside the markers are the steps between each two overlapped RF snippets.}
  \label{fig:time_complx}
\end{figure}

Since our RODNet starts with an M-Net chirp-merged input to the vanilla autoencoder with 3D convolution layers, further added with skip connections in hourglass (HG) structure and inception convolution layers, and eventually with TDC incorporation, it is important to know the performance influence of each functional module, as well as the computational complexities involved.

First of all, APs under different OLS thresholds are evaluated in Table~\ref{tab:results_apolss}. Here, we use different combinations of the backbones, supervision, and other modules in the RODNet.\footnote{Some experimental results shown in Table~\ref{tab:results_apolss} are also mentioned in our previous work \cite{wang2021rodnet}.} From the results in the table, the following conclusions can be reached: 1) The performance of the HG backbone is better than the vanilla backbone by around $5\%$. 2) Training the RODNet using CRF supervision can improve the performance by about $8\%$. 3) The customized modules, i.e., M-Net and TDC, along with temporal inception, can also each improve the detection performance by approximately $1\% - 2\%$, respectively.

Moreover, \textit{real-time} implementation is essential for autonomous or assisted driving applications. As mentioned in Section~\ref{subsec:postprocess}, we use different overlapping lengths of RF frames during the inference. With more overlapped frames, more robust detection results from the RODNet can be achieved, however, the inference time will also increase. 
The training and inference of the RODNet models are run on an NVIDIA Quadro GV100 GPU, and the time consumed is reported in Fig.~\ref{fig:time_complx}. 
Here, we show the AP of three building architectures (Vanilla, HG, and HG with temporal inception) for the RODNet, and use $100$ ms as a reasonable real-time threshold. 
The results illustrate that the RODNet with the relatively simpler vanilla backbone can achieve real-time and finish the prediction within $100$ ms. As for the HG backbone, it steps across the real-time threshold when the overlapping length increases. Moreover, HG without temporal inception layers is slightly faster than HG with all network components.

\begin{figure*}[t] 
\centering
\includegraphics[width=0.85\linewidth]{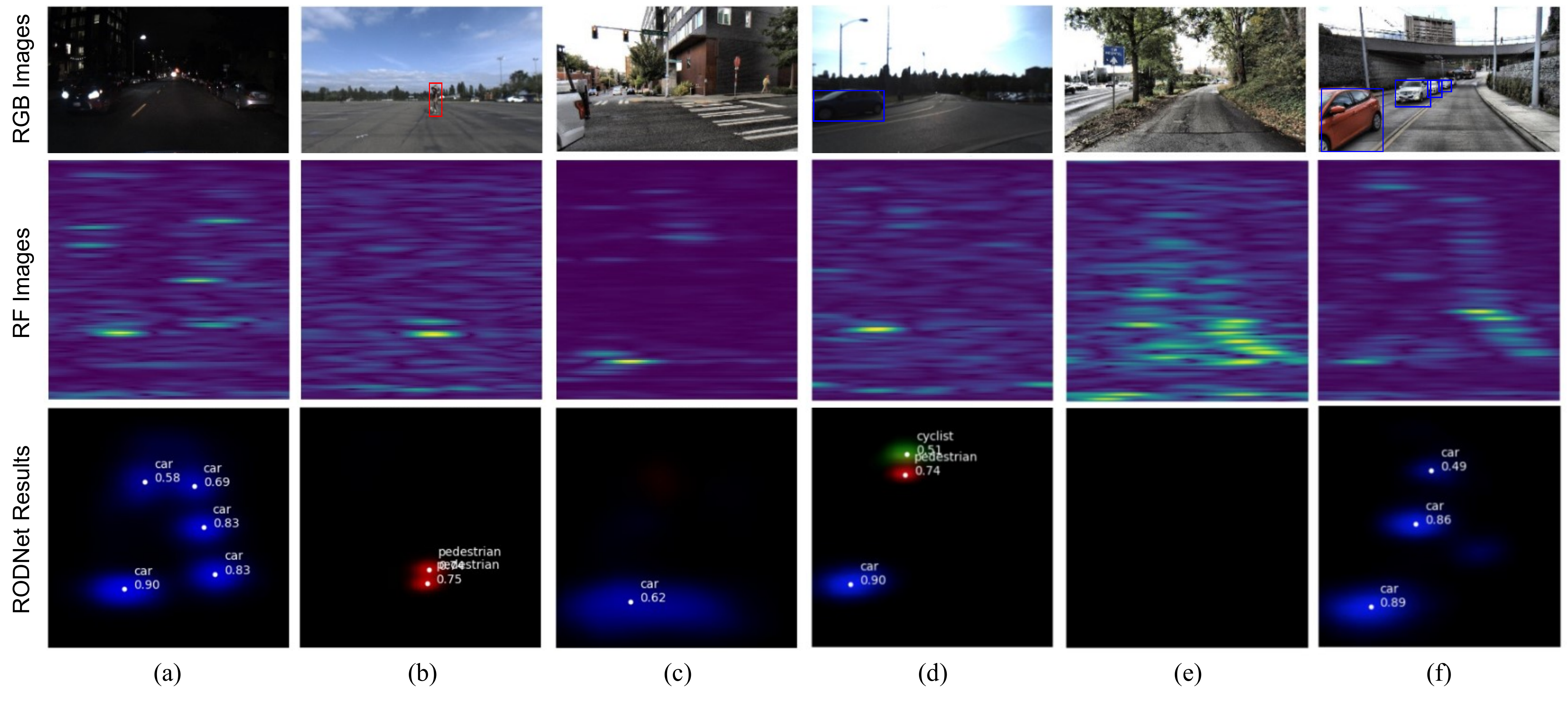}
  \caption{Examples illustrate the strengths of our RODNet. Conditions from left to right: (a) difficult for vision-based methods during the night; (b) visually occluded objects can be separately detected; (c) visually truncated vehicle can be detected; (d) pedestrian and cyclist fail to be detected under strong lighting condition; (e) \& (f) good detection results with noisy background.}
  \label{fig:res_stren_limit}
\end{figure*}

\subsubsection{RF Snippet Length}

\begin{wrapfigure}{r}{0.48\linewidth}
    \centering
    \includegraphics[width=\linewidth]{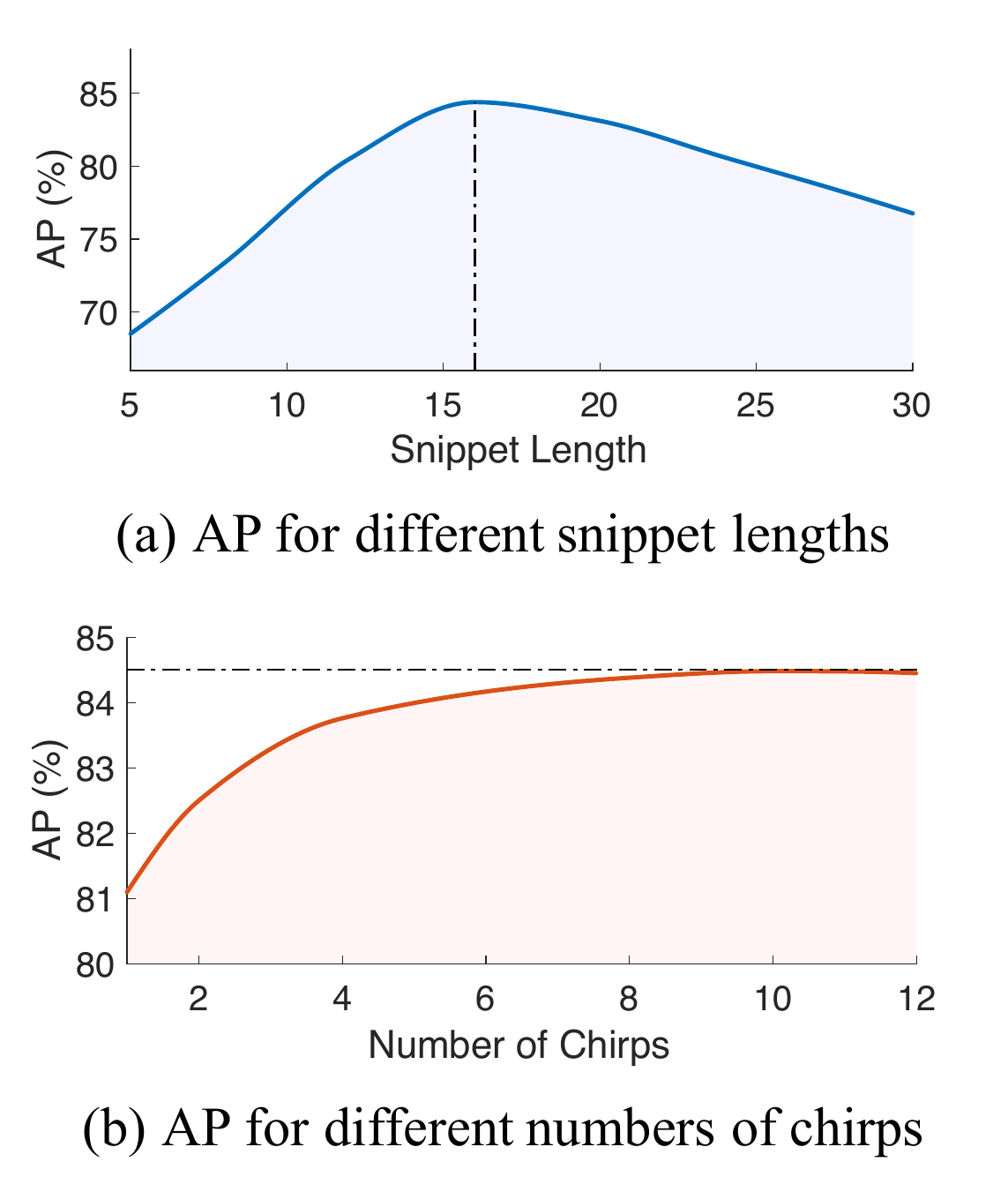}
    \caption{Results of different RF snippet length and number of chirps.}
    \label{fig:ablation_frame_chirp}
\end{wrapfigure}

Because our RODNet takes RF image snippets as the input, we would like to know how long the radar signals are required to obtain good detection results. 
Thus, we try different lengths of the snippets and evaluate their AP on our test set. The results are shown in Fig.~\ref{fig:ablation_frame_chirp}~(a). The experiments are operated on the backbone of HG (without temporal inception) with CRF, M-Net, and TDC. From the figure, AP is low with short input snippets because of insufficient radar information from a short temporal period. The detection AP of RODNet reaches the highest at length of $16$. While the snippet length continually increases, the AP starts to drop due to the snippet being too long to extract efficient features for radar object detection.

\subsubsection{Number of Chirps}

The number of chirps sent into the M-Net is another important parameter for our method. Larger number of chirps means more data are fed into the RODNet, resulting in higher input dimensions and time complexity. The experiments are also operated on the backbone of HG (without temporal inception) with CRF, M-Net and TDC. As shown in Fig.~\ref{fig:ablation_frame_chirp}~(b), the performance boosts with the number of chirps increases. However, it turns flat when the number of chirps is greater than $8$. Therefore, we choose $8$ to be the number of input chirps. Referring to the overall number of chirps $255$ per $\frac{1}{30}$-second RF frame used in the CRUW dataset, we use $8$ chirps (about $3\%$ of the radar data) to achieve favorable detection performance, which shows the efficiency of our method.

\subsubsection{Extracted Feature Visualization}

\begin{wrapfigure}{r}{0.4\linewidth}
\centering
\includegraphics[width=\linewidth]{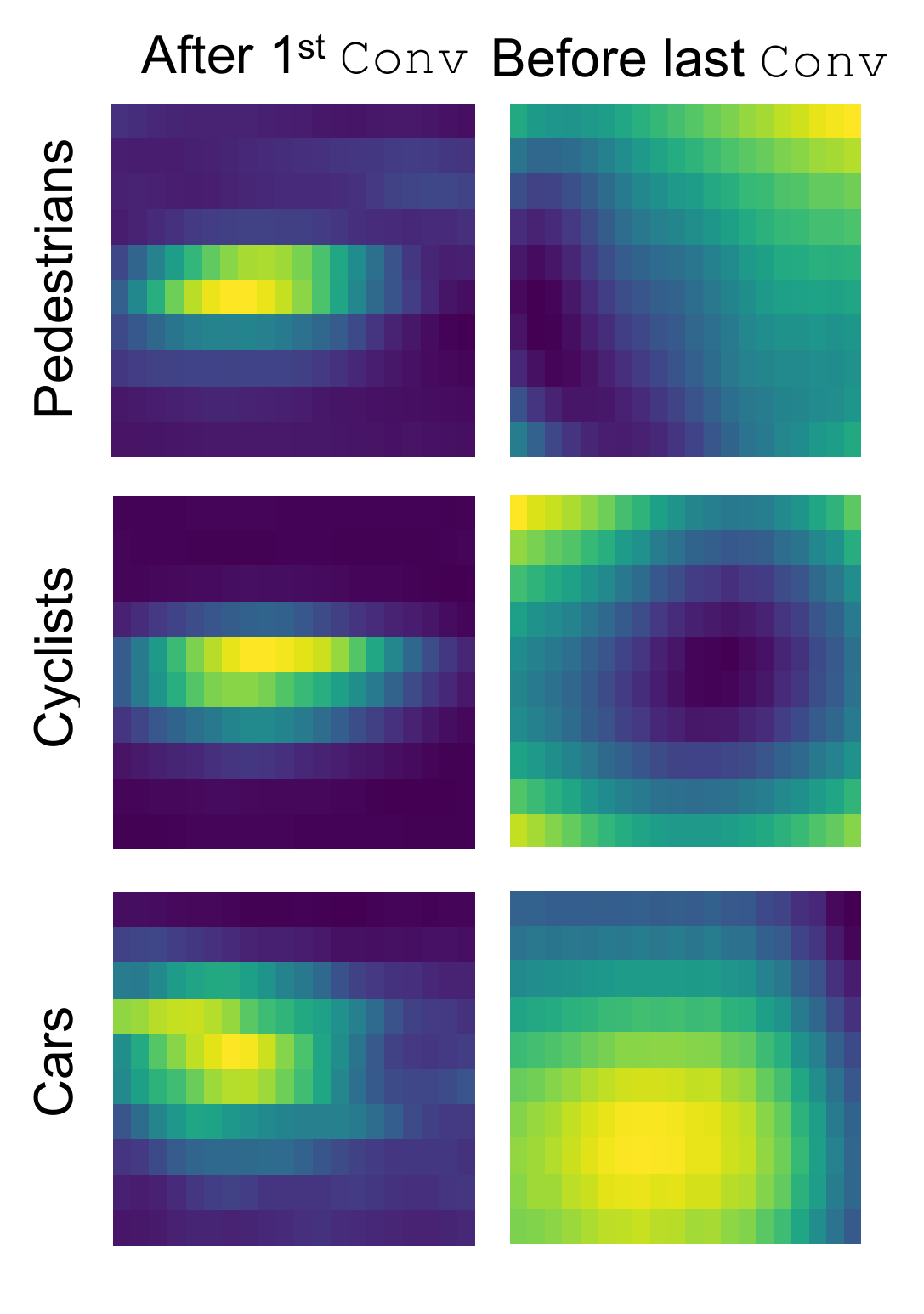}
  \caption{Feature visualization of different object classes. }
  \label{fig:feature_viz}
\end{wrapfigure}

After the RODNet is well-trained, we would like to analyze the features learned from the radar RF images. In Fig.~\ref{fig:feature_viz}, we show two different kinds of feature maps, i.e., the features after the first convolution layer and the features before the last layer of the RODNet. These feature maps are generated by cropping some randomly chosen objects from the original feature maps and average over all channels into one. From the visualization, we notice that the feature maps are similar in the beginning, and become significantly discriminative toward the end of the RODNet. Note that the visualized features are pixel-wise averaged within each object class to better represent the general class-level features.

\subsection{Strengths Comparing with Visual Object Detection}
\label{subsec:strength}

Some examples to illustrate the RODNet's advantages are shown in Fig.~\ref{fig:res_stren_limit}. First, the RODNet can maintain similar detection performance in different driving conditions, even during the night, as shown in the first example. Moreover, the RODNet can handle some occlusion cases when the camera can easily fail. In the second example, two pedestrians are nearly fully occluded in the image, but our RODNet can still separately detect both of them. This is because they are separated in the range of the radar point of view, where 3D information is revealed by the radar. 
Third, RF images have a wider FoV than RGB images so that the RODNet can see more information. As shown in the third example, only a small part of the car visible in the camera view, which can hardly be detected from the camera side, but the RODNet can successfully detect it. 
Besides, RODNet can detect objects that are not detected by image-based methods due to strong lighting condition. The fourth example shows that a pedestrian and a cyclist are detected by RODNet, but can hardly observed in the corresponding RGB image. Last but not least, our RODNet is able to distinguish noisy obstacles from objects after trained by CRF supervision. In the fifth and the last example, the noises, e.g., trees, walls, traffic signs, and poles, are suppressed in our RODNet results.

\section{Conclusion}
\label{sec:conclusion}

Object detection is crucial in autonomous driving and many other areas. Computer vision community has been focusing on this topic for decades and has come up with many good solutions. However, vision-based detection schemes are still suffering from many adverse lighting and weather conditions. This paper proposed a brand-new and novel object detection method purely from radar information, which can be more robust than vision in adverse conditions. The proposed RODNet can accurately and robustly detect objects, based on fully systematic cross-modal supervision scheme from an effective camera-radar fusion algorithm, in various autonomous and assisted driving scenarios even during the night or bad weather, which can potentially improve the role of radar in autonomous and assisted driving applications.


%


\appendices
\section{Network Architecture Implementation}
\label{appen:network_implement}

In this section, we describe the network architecture of the RODNet. We introduce two different kinds of backbones, i.e., vanilla and hourglass (HG). The detailed parameters of the neural network are presented in Table~\ref{tab:architecture_cdc} and Table~\ref{tab:architecture_hg}, where \texttt{Conv3D} denotes 3D convolution layer; \texttt{TDC} denotes temporal deformable convolution layer; \texttt{ConvT3D} denotes transpose 3D convolution layer; and \texttt{Conv3D(Skip)} denotes 3D convolution layer for skip connection.

\begin{table}[h]
\centering
\caption{RODNet with the vanilla backbone.}
\begin{tabular}{|cccc|}
\hline
Layer & Kernel & Stride & Channels  \\
\hline\hline
\texttt{Conv3D/TDC} & $(5,3,3)$ & $(1,1,1)$ & $64$ \\
\texttt{Conv3D/TDC} & $(5,3,3)$ & $(2,2,2)$ & $64$ \\
\texttt{Conv3D} & $(9,5,5)$ & $(1,1,1)$ & $128$ \\
\texttt{Conv3D} & $(9,5,5)$ & $(2,2,2)$ & $128$ \\
\texttt{Conv3D} & $(9,5,5)$ & $(1,1,1)$ & $256$ \\
\texttt{Conv3D} & $(9,5,5)$ & $(2,2,2)$ & $256$ \\
\hline\hline
\texttt{ConvT3D} & $(4,6,6)$ & $(2,2,2)$ & $128$ \\
\texttt{ConvT3D} & $(4,6,6)$ & $(2,2,2)$ & $64$ \\
\texttt{ConvT3D} & $(3,6,6)$ & $(1,2,2)$ & $3$ \\
\hline
\end{tabular}
\label{tab:architecture_cdc}
\end{table}

\begin{table}[h]
\centering
\caption{RODNet with the HG backbone (single stack).}
\begin{tabular}{|cccc|}
\hline
Layer & Kernel & Stride & Channels  \\
\hline\hline
\texttt{Conv3D/TDC} & $(5,3,3)$ & $(1,1,1)$ & $32$ \\
\texttt{Conv3D/TDC} & $(5,3,3)$ & $(1,1,1)$ & $64$ \\
\hline\hline
\texttt{Conv3D} & $(9,5,5)$ & $(1,1,1)$ & $64$ \\
\texttt{Conv3D} & $(9,5,5)$ & $(2,2,2)$ & $64$ \\
\texttt{Conv3D} & $(9,5,5)$ & $(1,1,1)$ & $128$ \\
\texttt{Conv3D} & $(9,5,5)$ & $(2,2,2)$ & $128$ \\
\texttt{Conv3D} & $(9,5,5)$ & $(1,1,1)$ & $256$ \\
\texttt{Conv3D} & $(9,5,5)$ & $(2,2,2)$ & $256$ \\
\hline\hline
\texttt{Conv3D(Skip)} & $(9,5,5)$ & $(1,1,1)$ & $64$ \\
\texttt{Conv3D(Skip)} & $(9,5,5)$ & $(2,2,2)$ & $64$ \\
\texttt{Conv3D(Skip)} & $(9,5,5)$ & $(1,1,1)$ & $128$ \\
\texttt{Conv3D(Skip)} & $(9,5,5)$ & $(2,2,2)$ & $128$ \\
\texttt{Conv3D(Skip)} & $(9,5,5)$ & $(1,1,1)$ & $256$ \\
\texttt{Conv3D(Skip)} & $(9,5,5)$ & $(2,2,2)$ & $256$ \\
\hline\hline
\texttt{ConvT3D} & $(4,6,6)$ & $(2,2,2)$ & $128$ \\
\texttt{ConvT3D} & $(4,6,6)$ & $(2,2,2)$ & $64$ \\
\texttt{ConvT3D} & $(3,6,6)$ & $(1,2,2)$ & $64$ \\
\texttt{Conv3D} & $(9,5,5)$ & $(1,1,1)$ & $3$ \\
\hline
\end{tabular}
\label{tab:architecture_hg}
\end{table}

\begin{figure}[h]
    \centering
    \includegraphics[width=0.85\linewidth]{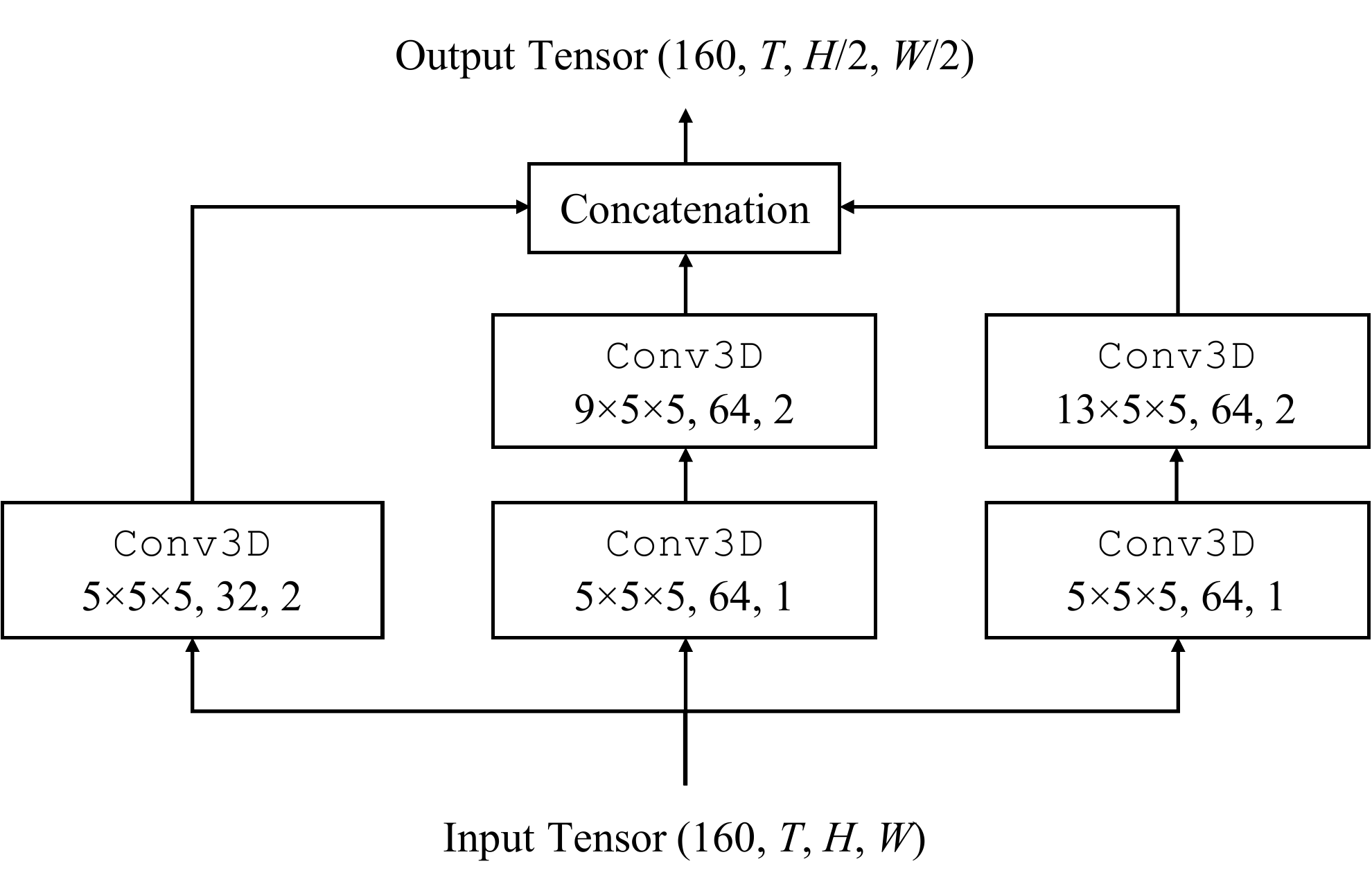}
    \caption{Illustration of the proposed temporal inception convolution layer. The three parameters inside \texttt{Conv3D} blocks are representing kernel size, number of channels, and spatial stride, respectively.}
    \label{fig:incep}
\end{figure}

The illustration of our proposed temporal inception convolution layer is shown in Fig.~\ref{fig:incep}. We utilize three different lengths for the temporal convolution kernels, i.e., $5$, $9$, $13$, to extract the features with different temporal lengths. The features of the three different temporal lengths are then concatenated in the channel domain and sent to the subsequent layers. For simplicity, we use $160$ as the number of both input and output channels of all temporal inception convolution layers. 

Moreover, we train the RODNet on an NVIDIA Quadro GV100 GPU. It takes about 4 days to train the RODNet (vanilla), 8 days to train the RODNet (HG), and 10 days to train the RODNet (HG) with temporal inception convolution layers.

\section{Implementation Details of Our Visual Teacher}

There are 5 modules in the teacher's pipeline. The implementation details are stated as follows:
\begin{itemize}
    \item Image-based object detection: pre-trained on the KITTI dataset and fine-tuned on the Cityscapes dataset.
    \item Depth estimation: a pre-trained self-supervised method that only needs stereo image pairs for training, and it is further retrained on our CRUW dataset.
    \item Semantic segmentation: pre-trained on the Cityscapes dataset.
    \item Multi-object tracking: pre-trained on the KITTI dataset and can be easily generalized to our CRUW dataset without much performance degradation.
\end{itemize}

\section{Temporal Deformable Convolution Back-propagation}
\label{appen:tdc_back}

According to the temporal deformable convolution defined in Eq.~4 (in the original manuscript), the gradient w.r.t. the offset field $\Delta \mathbf{p}_n$ can be calculated as
\begin{equation}
\begin{aligned}
    &\frac{\partial\mathbf{y}(\mathbf{p}_0)}{\partial\Delta \mathbf{p}_n} 
    = \sum_{\mathbf{p}_n \in \mathcal{R}} \mathbf{w}(\mathbf{p}_n) \cdot \frac{\partial\mathbf{x}(\mathbf{p}_0 + \mathbf{p}_n + \Delta \mathbf{p}_n)}{\partial\Delta \mathbf{p}_n}\\
    &= \sum_{\mathbf{p}_n \in \mathcal{R}} \left[ \mathbf{w}(\mathbf{p}_n) \cdot \sum_{\mathbf{q}} \frac{\partial G(\mathbf{q}, \mathbf{p}_0 + \mathbf{p}_n + \Delta \mathbf{p}_n)}{\partial\Delta \mathbf{p}_n} \mathbf{x}(\mathbf{q}) \right].
\end{aligned}
\end{equation}
Here, $\mathbf{q}$ enumerates all integer locations in the 3D feature map $\mathbf{x}$, and $G(\cdot,\cdot)$ is the bilinear interpolation kernel on the 2D offset field defined as
\begin{equation}
    G(\mathbf{q}, \mathbf{p}) = \left[ g(q_u, p_u), g(q_v, p_v) \right],
\end{equation}
where $g(a, b) = \max (0, 1-|a-b|)$.

\section{Additional Qualitative Results}

Some additional qualitative results are shown in Fig.~\ref{fig:res_qual_vis_supp}, where different colors represent different detected object classes (red: pedestrian; green: cyclist; blue: car). Based on the current hardware capability, the radar's FoV is $0$-$25$m, $\pm90^{\circ}$. Our RODNet presents promising radar object detection performance in various driving scenarios, i.e., parking lot, campus road, city street, with different lighting conditions. More qualitative results can be found in our supplementary demo video. 

\begin{figure*}[t]
    \centering
    \includegraphics[width=0.9\linewidth]{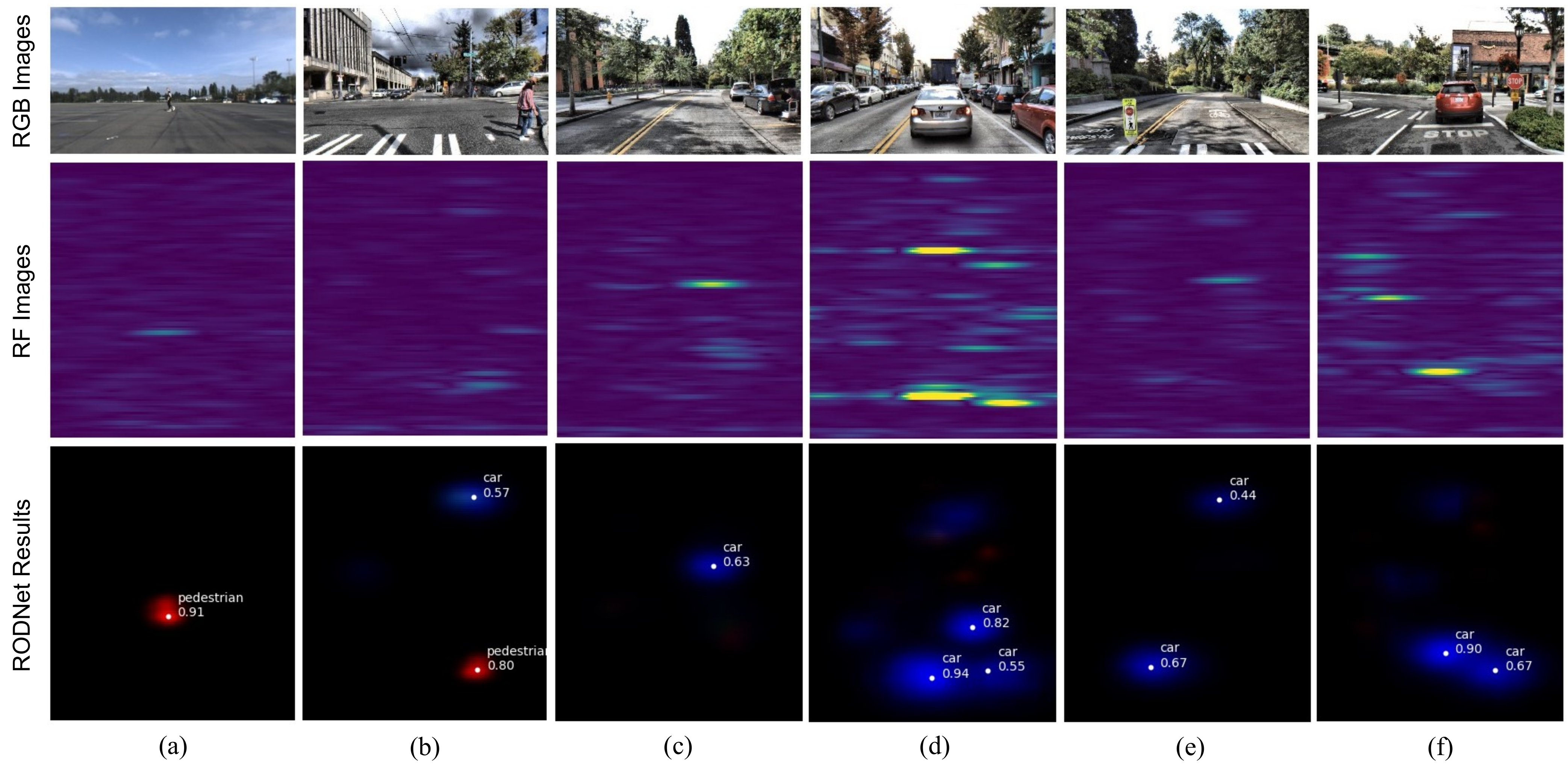}
    \caption{The failure cases of our RODNet. (a) \& (b): False negatives when multiple pedestrians are very near (two pedestrians in both scenes); (c): False negative for a car whose trunk is open (abnormal surface feature for a car); (d): False negatives due to the strong reflection of the nearby object in the static scenario; (e) \& (f): False positives from traffic signs.}
\label{fig:failure}
\end{figure*}

Besides, we also show some failure cases of our RODNet in Fig.~\ref{fig:failure}. In Fig.~\ref{fig:failure}, (a) and (b) show some false negatives of pedestrian detection. Since pedestrians usually have a much smaller volume than cars, so they cannot be easily detected separately when they are next to each other. Fig.~\ref{fig:failure} (c) and (d) are two examples of false negatives of car detections, where (c) shows a car with the back trunk opened is missing in our RODNet detection because of abnormal surface features, and (d) shows that the reflection signals from the faraway cars are suppressed by a nearby car in the static scenario. Finally, Fig.~\ref{fig:failure} (e) and (f) show some false positives caused by traffic signs.

\begin{figure*}[t]
    \centering
    \includegraphics[width=0.9\linewidth]{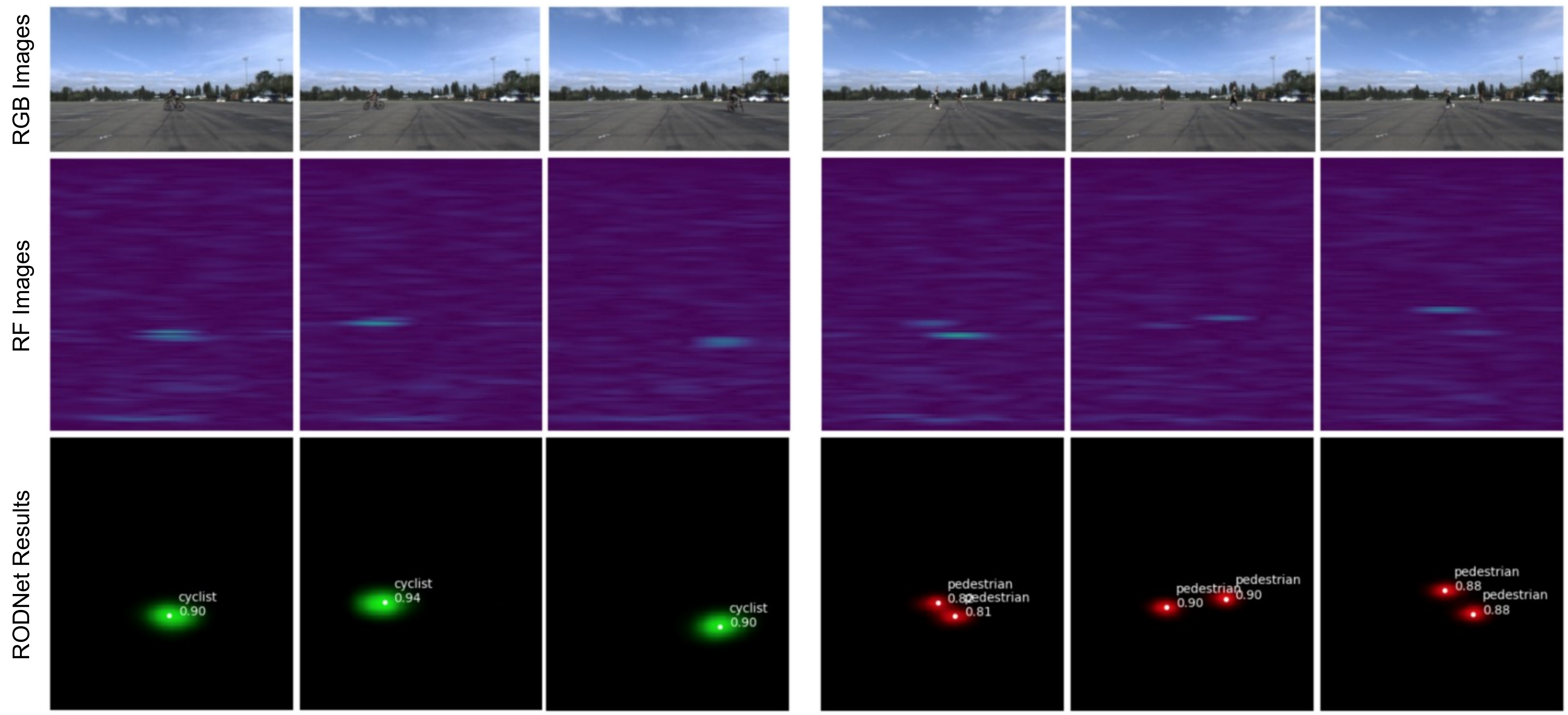}\vspace{0.5em}
    \includegraphics[width=0.9\linewidth]{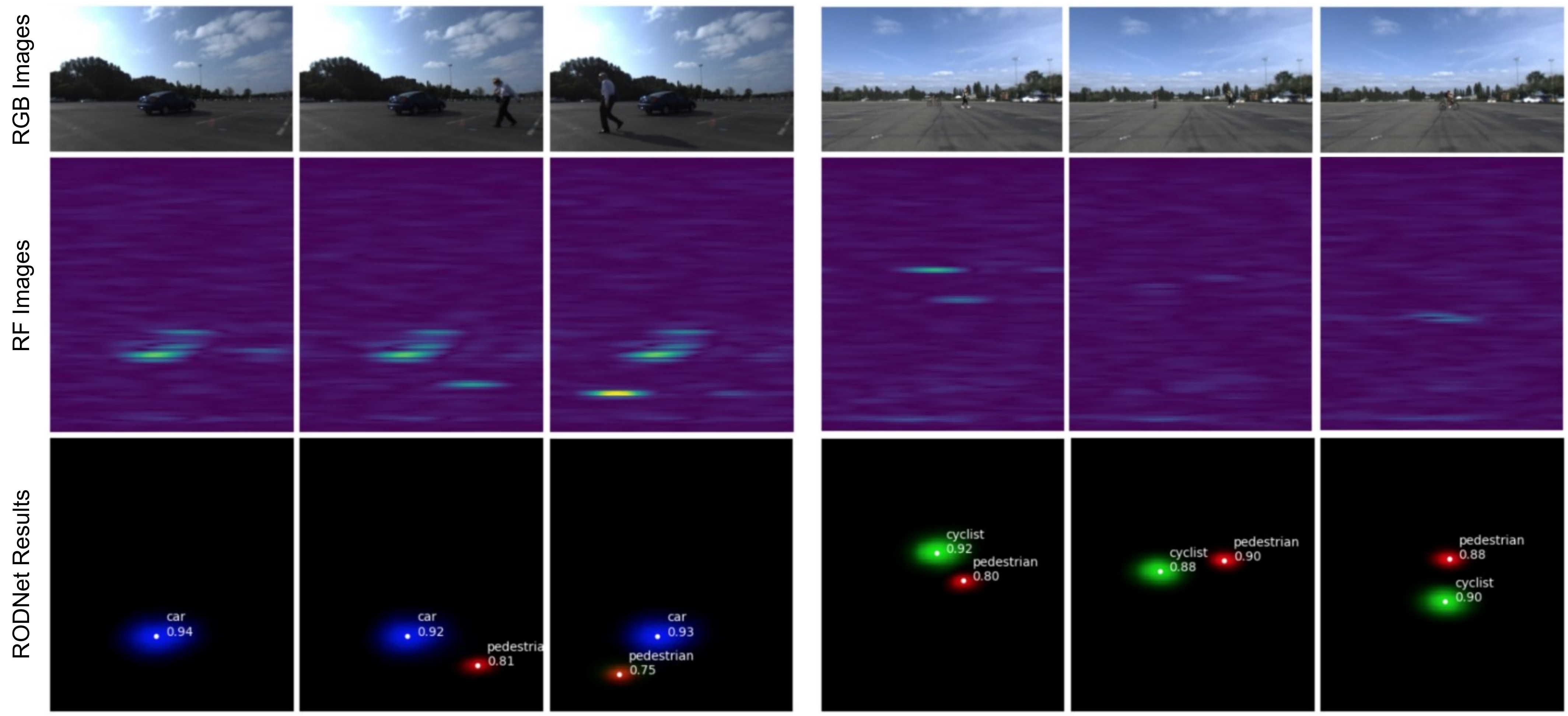}\vspace{0.5em}
    \includegraphics[width=0.9\linewidth]{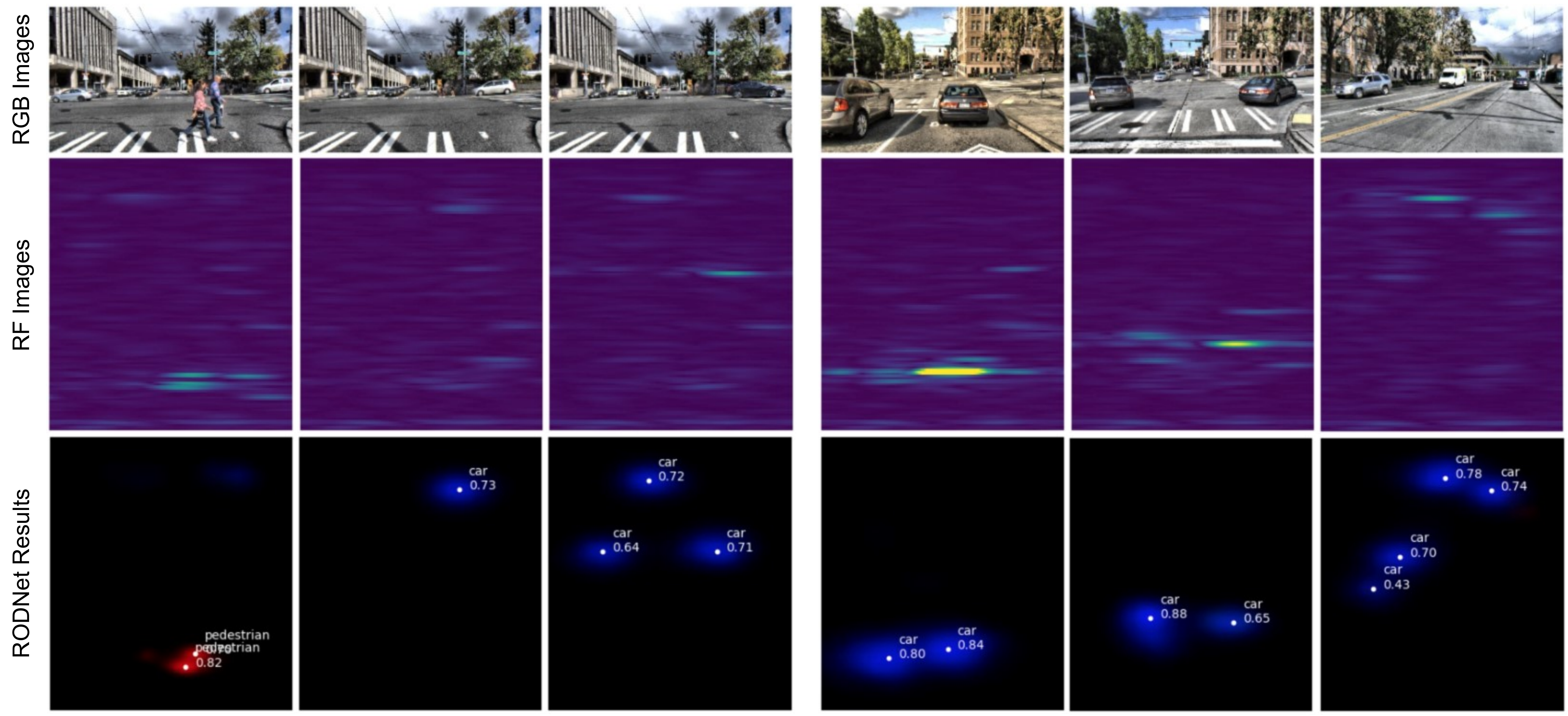}
    \caption{The qualitative results of the RODNet in different driving scenarios (12 sequences in total). The three rows are RGB images, RF images, and resulting ConfMaps with final detections, respectively. The white dots on the ConfMaps represent the final detections after post-processing.}
\end{figure*}

\begin{figure*}[t]\ContinuedFloat
    \centering
    \includegraphics[width=0.9\linewidth]{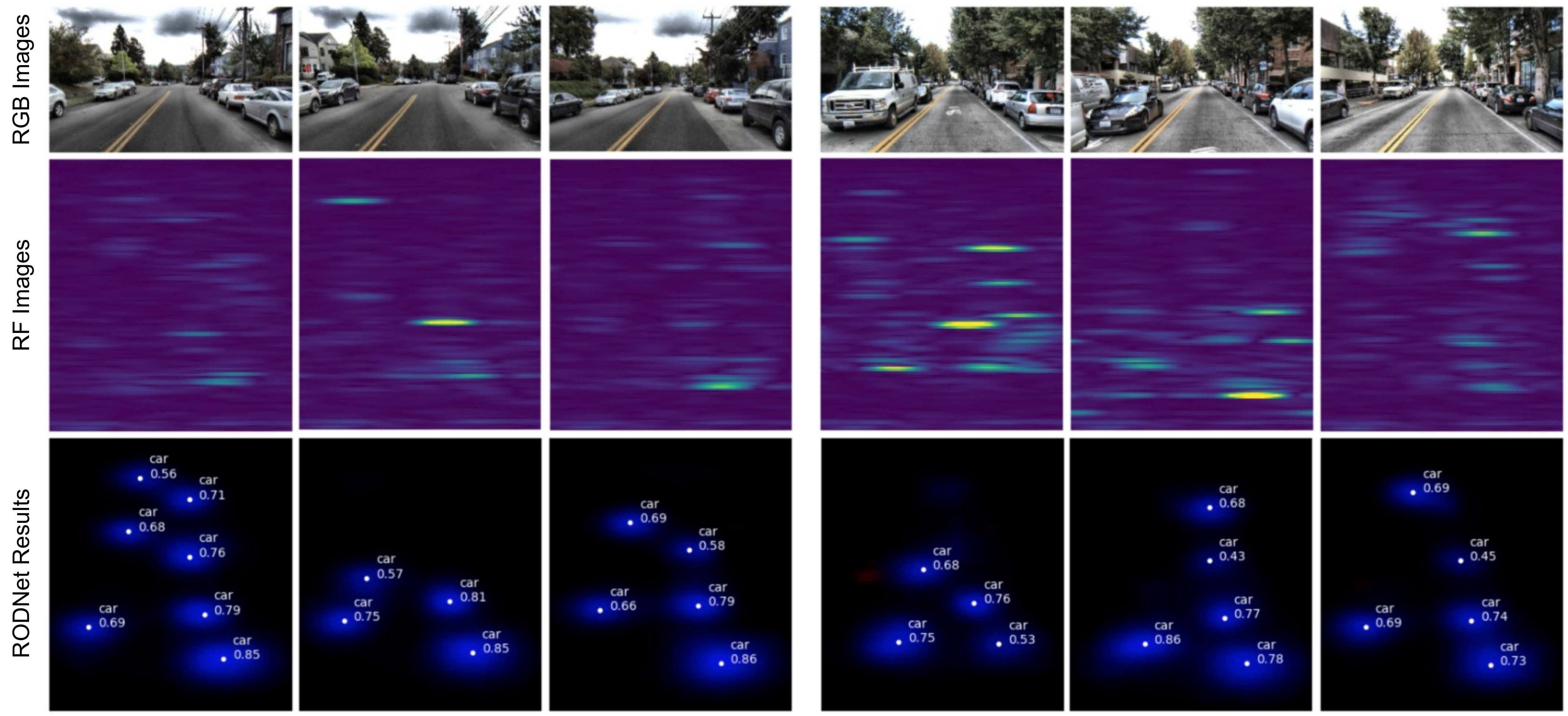}\vspace{0.5em}
    \includegraphics[width=0.9\linewidth]{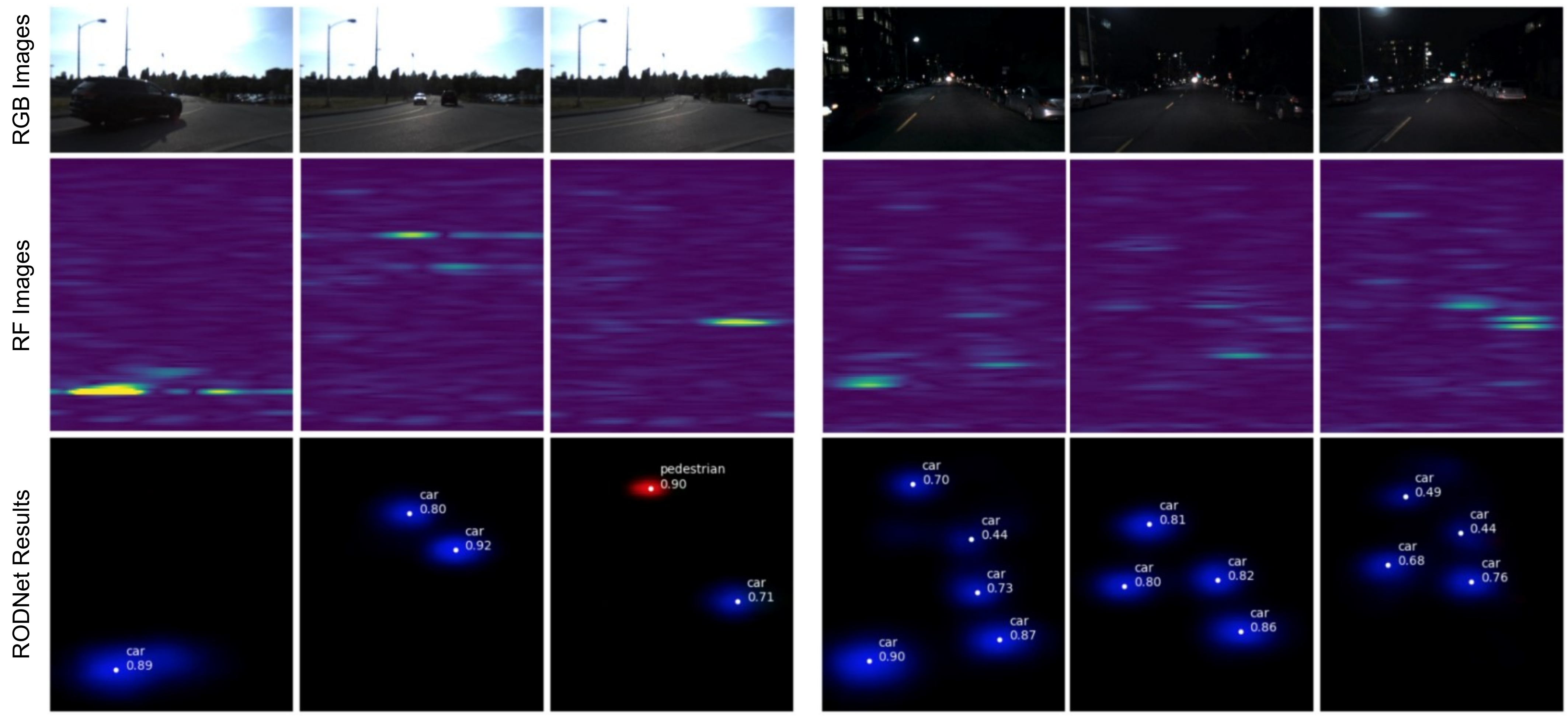}\vspace{0.5em}
    \includegraphics[width=0.9\linewidth]{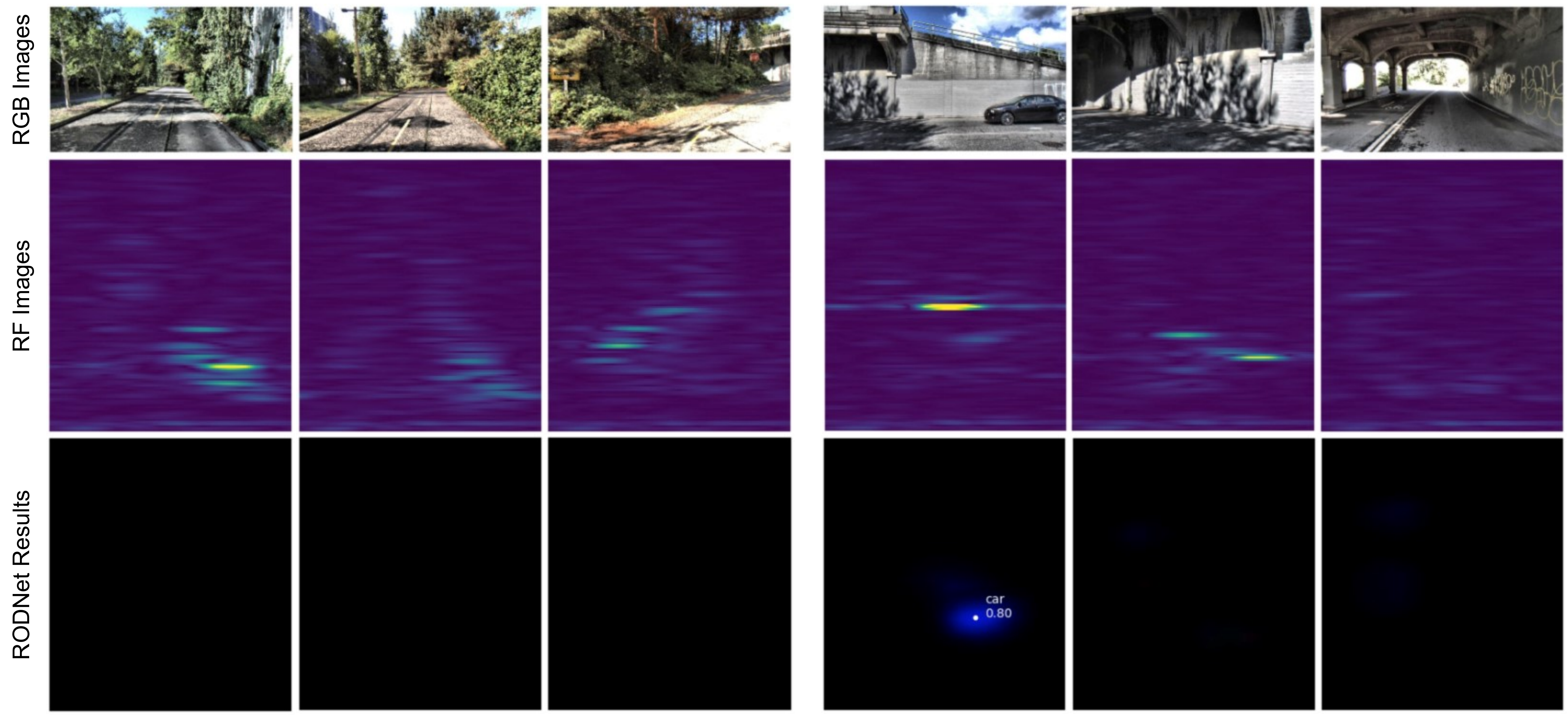}
    \caption{The qualitative results of the RODNet in different driving scenarios (12 sequences in total). The three rows are RGB images, RF images, and resulting ConfMaps with final detections, respectively. The white dots on the ConfMaps represent the final detections after post-processing. (Continued)}
\label{fig:res_qual_vis_supp}
\end{figure*}

\section*{Acknowledgment}

This work was partially supported by CMMB Vision -- UWECE Center on Satellite Multimedia and Connected Vehicles. The authors would also like to thank the colleagues and students in Information Processing Lab at UWECE for their help and assistance on the dataset collection, processing, and annotation works.

\ifCLASSOPTIONcaptionsoff
  \newpage
\fi



\bibliographystyle{IEEEtran}
\bibliography{IEEEabrv,IEEEtran}
%
%
%

%

\begin{IEEEbiography}[{\includegraphics[width=1in,height=1.25in,clip,keepaspectratio]{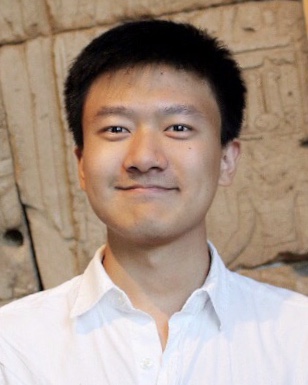}}]{Yizhou Wang}
is a Ph.D. student in Electrical and Computer Engineering at the University of Washington, advised by Prof. Jenq-Neng Hwang. He received his M.S. in Electrical Engineering in 2018 from Columbia University, advised by Prof. Shih-Fu Chang. His research interests include autonomous driving, computer vision, deep learning, and cross-modal learning. 
\end{IEEEbiography}

\begin{IEEEbiography}[{\includegraphics[width=1in,height=1.25in,clip,keepaspectratio]{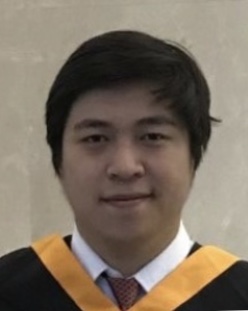}}]{Zhongyu Jiang}
received the B.E. degree in Computer Science and Technology from Tsinghua University, Beijing, China, in 2018 and M.Sc. degree in Computer Science and System from the University of Washington, Tacoma. He is currently a Ph.D. student in Electrical and Computer Engineering, University of Washington.
\end{IEEEbiography}

\begin{IEEEbiography}[{\includegraphics[width=1in,height=1.3in,clip,keepaspectratio]{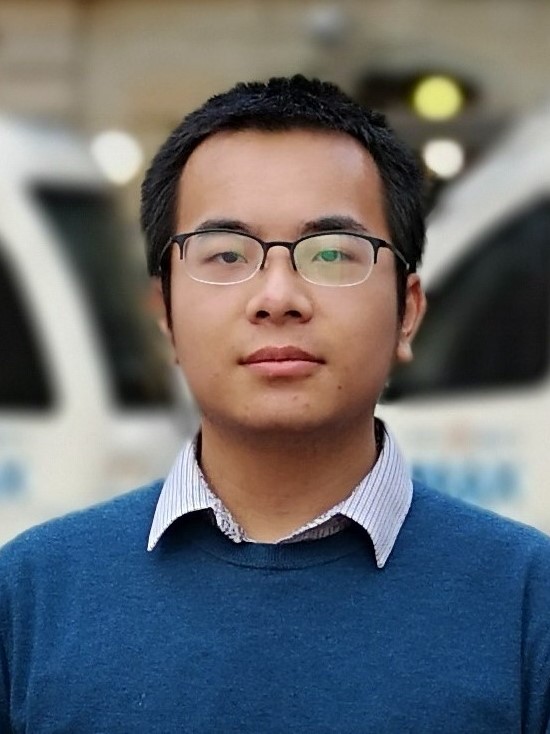}}]{Yudong Li} 
is an undergraduate junior in the Information School at the University of Washington. He is pursuing the B.S. degree in Informatics, and planning on graduation in 2022. Currently, he is a research assistant at the Information Processing Lab, University of Washington, facilitating the works on machine learning related topics.
\end{IEEEbiography}

\begin{IEEEbiography}[{\includegraphics[width=1in,height=1.5in,clip,keepaspectratio]{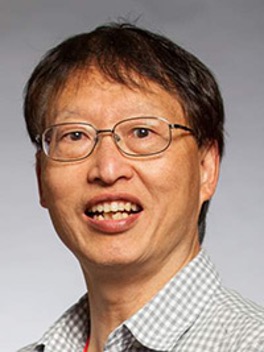}}]{Jenq-Neng Hwang}
received the BS and MS degrees, both in electrical engineering from the National Taiwan University, Taipei, Taiwan, in 1981 and 1983 separately. He then received his Ph.D. degree from the University of Southern California. In the summer of 1989, Dr. Hwang joined the Department of Electrical and Computer Engineering (ECE) of the University of Washington in Seattle, where he has been promoted to Full Professor since 1999. He served as the Associate Chair for Research from 2003 to 2005, and from 2011-2015. He also served as the Associate Chair for Global Affairs from 2015-2020. He is the founder and co-director of the Information Processing Lab, which has won CVPR AI City Challenges awards in the past years. He has written more than 380 journal, conference papers and book chapters in the areas of machine learning, multimedia signal processing, and multimedia system integration and networking, including an authored textbook on ``Multimedia Networking: from Theory to Practice'', published by Cambridge University Press. Dr. Hwang has close working relationship with the industry on multimedia signal processing and multimedia networking. 
Dr. Hwang received the 1995 IEEE Signal Processing Society's Best Journal Paper Award. He is a founding member of Multimedia Signal Processing Technical Committee of IEEE Signal Processing Society and was the Society's representative to IEEE Neural Network Council from 1996 to 2000. He is currently a member of Multimedia Technical Committee (MMTC) of IEEE Communication Society and also a member of Multimedia Signal Processing Technical Committee (MMSP TC) of IEEE Signal Processing Society. He served as associate editors for IEEE T-SP, T-NN and T-CSVT, T-IP and Signal Processing Magazine (SPM). He is currently on the editorial board of ZTE Communications, ETRI, IJDMB and JSPS journals. He served as the Program Co-Chair of IEEE ICME 2016 and was the Program Co-Chairs of ICASSP 1998 and ISCAS 2009. Dr. Hwang is a fellow of IEEE since 2001.
\end{IEEEbiography}

\begin{IEEEbiography}[{\includegraphics[width=1in,height=1.3in,clip,keepaspectratio]{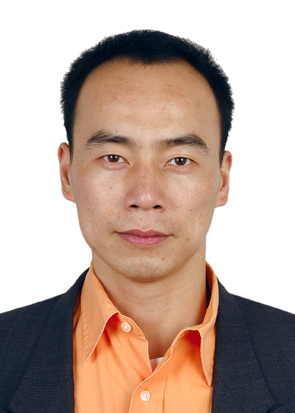}}]{Guanbin Xing} 
received his B.S. and M.S. degrees from Peking University, Beijing, China, in 1996 and 1999, respectively, and his Ph.D. degree from the University of Washington in 2004, all in electrical engineering. 
Dr. Xing has over 15 years of experience as a senior system architect in the Wireless Communications and Digital Broadcasting industry. In 2017, he joined CMMB Vision - UW EE Center on Satellite Multimedia and Connected Vehicles as a research scientist working on the mmWave radar signal processing and machine learning based sensor fusion solutions for autonomous driving.
\end{IEEEbiography}

\begin{IEEEbiography}[{\includegraphics[width=1in,height=1.3in,clip,keepaspectratio]{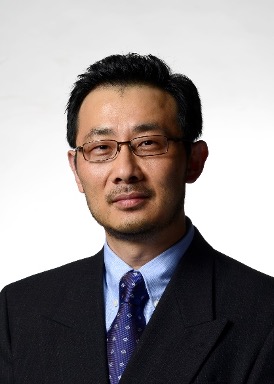}}]{Hui Liu}
is the President and CTO of Silkwave Holdings and an affiliate professor at University of Washington. He received his B.S. in 1988 from Fudan University, Shanghai, China, and a Ph.D. degree in 1995 from the Univ. of Texas at Austin, all in electrical engineering. He was previously a full professor/associate chair at the Dept. of EE, Univ. of Washington and a chair professor/associate dean of School of Electronic, Information \& Electrical Engineering (SEIEE) at Shanghai Jiao Tong University. Dr. Liu was one of the principal designers of the 3G TD-SCDMA mobile technologies, and the founder of Adaptix which pioneered the development of OFDMA-based mobile broadband networks (mobile WiMAX and 4G LTE). His research interests include broadband wireless networks, satellite communications, digital broadcasting, machine learning and autonomous driving. 
Dr. Liu has published more than 90 journal articles, 2 textbooks, and has over 80 awarded patents. He was selected Fellow of IEEE for contributions to global standards for broadband cellular and mobile broadcasting. He is the General Chairman for the 2005 Asilomar conference on Signals, Systems, and Computers and the 2014 IEEE/CIC International Conference on Communications in China (ICCC14). He is a recipient of 1997 National Science Foundation (NSF) CAREER Award, the Gold Prize Patent Award in China, 3 IEEE best conference paper awards, and the 2000 Office of Naval Research (ONR) Young Investigator Award. 
\end{IEEEbiography}






\end{document}